\documentclass[runningheads]{llncs}
\usepackage[utf8]{inputenc}
\usepackage{tabularx}
\newcolumntype{Y}{>{\centering\arraybackslash}X}
\usepackage{subcaption}
\usepackage{multirow}
\usepackage{graphicx}
\usepackage{xcolor, colortbl}
\usepackage{amsmath, mathrsfs}
\usepackage{amssymb}
\usepackage{paralist}
\usepackage{url}
\usepackage{enumitem}
\usepackage{fancyvrb}
\usepackage{algorithm}
\usepackage[backref=page]{hyperref}
\usepackage{algpseudocode}

\newcommand{\furl}[1]{\footnote{\scriptsize \url{#1}}}

\usepackage{forest}
\begin{document}
\title{Scaling Up Knowledge Graph Creation to Large and Heterogeneous Data Sources}
%\titlerunning{}
\author{Enrique Iglesias\inst{1}
Samaneh Jozashoori\inst{2,3}, 
        Maria-Esther Vidal\inst{1,2,3}}
%
%\authorrunning{}
\institute{L3S Research Center, Leibniz University of Hannover, Germany \\ 
\email{iglesias@l3s.de}\\
\and
TIB Leibniz Information Center for Science and Technology, Germany\\
\email{samaneh.jozashoori,maria.vidal@tib.eu}
\and
Leibniz University of Hannover
}
\maketitle         % typeset the header of the contribution
%
%%%%%%%%%%%%%%%%%%%%%%%%%%%%%%%%%%%%%%%%
%%              ABSTRACT              %%
%%%%%%%%%%%%%%%%%%%%%%%%%%%%%%%%%%%%%%%%

\begin{abstract}
RDF knowledge graphs (KG) are powerful data structures to represent factual statements created from heterogeneous data sources. KG creation is laborious and demands data management techniques to be executed efficiently. This paper tackles the problem of the automatic generation of KG creation processes declaratively specified; it proposes techniques for planning and transforming heterogeneous data into RDF triples following mapping assertions specified in the RDF Mapping Language (RML). Given a set of mapping assertions, the planner provides an optimized execution plan by partitioning and scheduling the execution of the assertions. First, the planner assesses an optimized number of partitions considering the number of data sources, type of mapping assertions, and the associations between different assertions. After providing a list of partitions and assertions that belong to each partition, the planner determines their execution order. A greedy algorithm is implemented to generate the partitions' bushy tree execution plan. Bushy tree plans are translated into operating system commands that guide the execution of the partitions of the mapping assertions in the order indicated by the bushy tree. The proposed optimization approach is evaluated over state-of-the-art RML-compliant engines, and existing benchmarks of data sources and RML triples maps. Our experimental results suggest that the performance of the studied engines can be considerably improved, particularly in a complex setting with numerous triples maps and large data sources. As a result, engines that time out in complex cases are enabled to produce at least a portion of the KG applying the planner.

\keywords{Knowledge graph creation, Data Integration Systems, RDF Mapping Languages, Query Execution Planning}
\end{abstract}

%%%%%%%%%%%%%%%%%%%%%%%%%%%%%%%%%%%%%%%%
%%            INTRODUCTION            %%
%%%%%%%%%%%%%%%%%%%%%%%%%%%%%%%%%%%%%%%%
\section{Introduction}
Knowledge graphs (KGs) are data structures able to converge data and metadata collected from various data sources \cite{gutierrez2021knowledge}. Data sources can be heterogeneous and comprise structured, semi-structured, or unstructured data. Nevertheless, several parameters like number and type of mapping assertions and data source complexities like large volume, heterogeneity, and high duplicate rates may considerably affect the performance of KG creation. 
\\
\noindent The declarative definition of KGs using W3C standard languages like R2RML~\cite{das2012r2rml} and RML~\cite{DimouSCVMW14} have gained momentum, and numerous real-world applications (e.g., \cite{abs-2012-01953,en15113973,KritharaARNBVMG19}) resort to these formalisms to provide transparent, maintainable, and traceable processes of KG creation. As a result, the scientific community has actively contributed to the problem of KG creation with formal systems to formalize the whole process~\cite{DBLP:journals/semweb/KejriwalSL19,namici2018comparing,DBLP:conf/www/PriyatnaCS14}, theoretical and empirical analyses of parameters that affect the process performance~\cite{Chaves-FragaEIC19,lenzerini2002data}, engines for KG creation~\cite{Chaves-FragaRPV21,DBLP:conf/www/DimouNVMW16,iglesias2020sdm,rocketrml}, and benchmarks to assess the engines' performance~\cite{chaves2020gtfs}. Despite these recent advances, existing engines could still struggle to perform well in real-world settings. For example,  KG creation in biomedicine demands the integration of various data types~\cite{VidalEJSR19}, e.g., genes, drugs, scientific publications, and clinical records, which change frequently. Complex pipelines composed of numerous mapping rules (e.g., more than 1,000 rules) collecting data from sources in a myriad of formats (e.g., relational or textual) may be costly in terms of time and memory consumption. 
\\
\noindent
Our work is inspired by our experience in developing such complex pipelines in the context of the EU funded projects iASiS~\cite{iasis}, BigMedilytics~\cite{bm}, and CLARIFY~\cite{clarify}, as well as in  CoyPu~\cite{coypu}, a German project funded by the Federal Ministry of Economics and Climate Protection~\cite{bmwk}. Specifically, the SDM-Genomic benchmark~\cite{sdm-genomic,sdm-genomic-mapping} is inspired by the computational challenges addressed during the integration of genomic data from the COSMIC database~\cite{cosmic} into the KGs of the biomedical projects. These mapping assertions are complex regarding dataset size, the number of mapping assertions, and types of joins among them. Initially, none of the existing engines (e.g., RMLMapper~\cite{dimou_ldow_2016}, RocketRML~\cite{rocketrml}, and SDM-RDFizer~\cite{iglesias2020sdm}) was able to run the complex mapping assertions on the project data in a reasonable time (e.g., less than 48 hours). Since biomedical data change frequently, these mappings are executed periodically. Manually, knowledge engineers rewrote the mapping assertions~\cite{genomic} and transformed them into simpler rules executable by the SDM-RDFizer. These transformations inspired the proposed optimization techniques for mapping assertions.  
%Accordingly, KG creation engines should be able to plan and optimize the execution of mapping assertions to minimize the execution cost. However, given the complexity of the KG creation process, scalable approaches still represent a challenge to data management.  

\noindent \textbf{Problem Statement and Objectives.}
We tackle the problem of efficiently executing KG creation when the process is declaratively defined using mapping languages like R2RML or RML (a.k.a. [R2]RML).
We formalize the problem as an optimization problem, where mapping assertions are grouped and scheduled into execution plans that reduce execution time or memory consumption. A solution to the problem is an execution plan of groups of mapping assertions scheduled as a binary bushy tree~\cite{DBLP:conf/pods/ScheufeleM97}; this execution avoids the sequential execution of the mapping assertions and reduces the complexity of duplicate removal. The problem of generating such execution plans is known to be NP-hard~\cite{DBLP:conf/pods/ScheufeleM97} in general. Thus, our objective is to efficiently traverse the space of execution plans and generate a plan that scales up to complex scenarios. 

\noindent \textbf{Our Proposed Solution.}
We propose a heuristic-based approach that groups mapping assertions executed against at most two data sources. The execution of the identified groups of mapping assertions is scheduled in a bushy tree, where duplicate removal is executed as soon as possible, i.e., they are pushed down into the tree and executed following an eager evaluation approach. We present two greedy approaches; one algorithm partitions the mapping assertions into groups, while the other generates bushy trees that schedule the groups' execution. The approach is \emph{engine agnostic}, i.e., the execution plan can be executed in any of the existing KG creation engines to speed up the KG creation process.
Empirically, we study the performance of the proposed approach and the generated plans. The study assesses the performance of state-of-the-art RML engines on existing benchmarks of KG creation. The observed outcomes put in perspective the benefits of scheduling the execution of mapping assertions following the generated plans. Moreover, these results indicate that not only can the process of KG creation be accelerated, but also consumed memory is reduced. 
\noindent \textbf{Contributions.} 
In summary, the scientific contributions of this work are as follows: 
\begin{itemize}[noitemsep]
\item\textbf{Engine-Agnostic Execution Planning Techniques for Knowledge Graph Creation}. We formalize the KG creation process and present greedy algorithms to generate execution plans that enable the efficient execution of KG creation pipelines. The proposed execution planning techniques implement a two-fold approach. First, mapping assertions are partitioned to avoid more than one join between two different mapping assertions executed in one group. Then, groups of mapping assertions are combined greedily to ensure those that generate instances of the same overlapped predicates are placed lower in the tree to be executed as soon as possible.
%these groups are connected by an internal tree node corresponding to the duplicate removal. 
%The execution of duplicate removal requires both groups of mapping assertions to be completely executed before starting the removal. Contrary, groups of mapping assertions with no overlapping predicates can be executed in parallel, and the results can be merged incrementally. 
\item\textbf{Execution Methods for Knowledge Graph Creation}.
We propose engine-agnostic techniques for the execution of mapping assertions. They translate a bushy tree plan into operating system commands to execute mapping assertions following the order indicated in the bushy tree plan. In case of duplicated RDF triples generated by the execution of groups of assertions, duplicate removal operators are scheduled and executed as soon as possible. This strategy reduces execution time and memory consumption and enables continuous generation of RDF triples. 
\begin{figure*}[t!]
\begin{center}

    \includegraphics[width=1.0\textwidth]{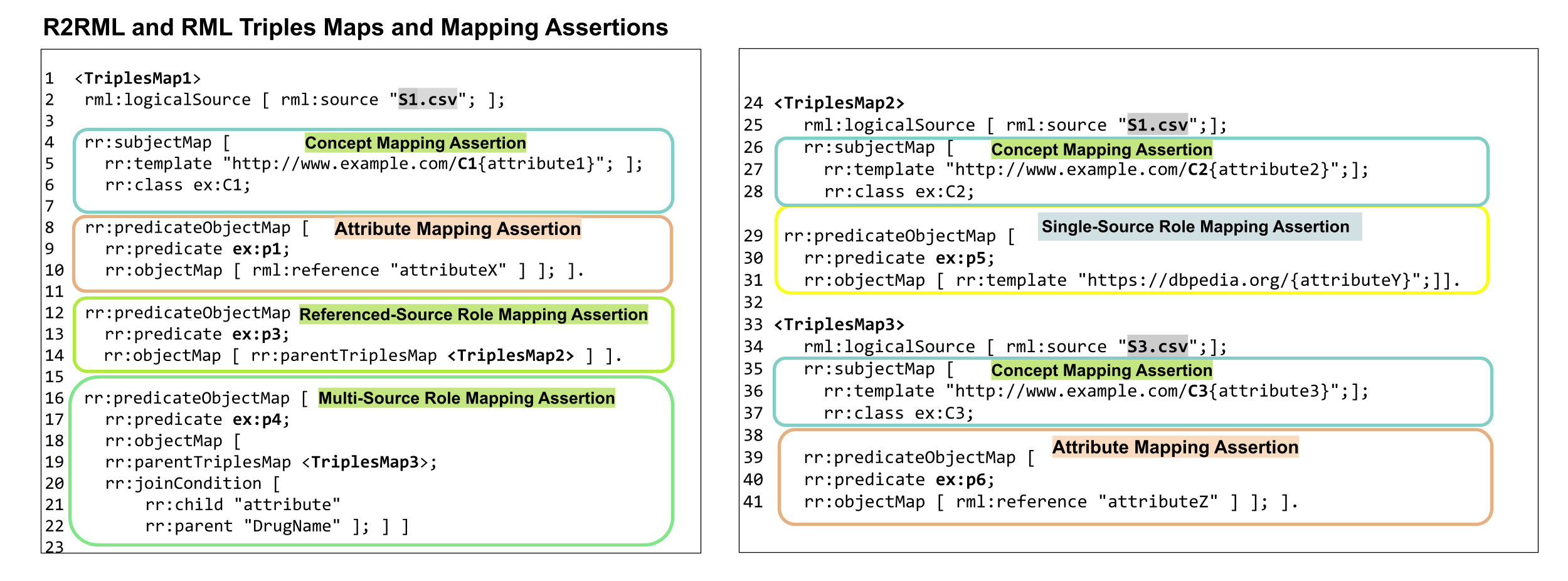}
    \caption{\textbf{Mapping Assertions.} Mapping assertions are expressed in R2RML --the W3C recommendation standard-- and its extension RML. The example comprises a) three concept mapping assertions defining the classes C1, C2, and C3; b) two attribute mapping assertions for the definition of  attributeX and attributeY, and c) two role mapping assertions: one referencing assertions defined over the same logical source (i.e., referenced-source), and the other one, referencing assertions defined over different sources (i.e., multi-source). }
         \label{fig:preliminaries}  
\end{center}
\end{figure*}
\item\textbf{Experimental Assessment of the Proposed  Methods}.
 We report on the empirical evaluation of the proposed methods in two benchmarks, SDM-Genomic-Datasets~\cite{sdm-genomic} and the GTFS-Madrid-Bench~\cite{chaves2020gtfs}, and  four  [R2]RML-compliant engines: RMLMapper~\cite{DimouSCVMW14}, RocketRML~\cite{rocketrml}, Morph-KGC~\cite{arenas2022morph}, and SDM-RDFizer~\cite{iglesias2020sdm}. In total, 236 testbeds are executed and analyzed. These results suggest savings in execution time of up to 76.09\%. Moreover, the proposed execution planning techniques enable the incremental generation of RDF triples. Thus, engines like RMLMapper which times out after five hours with zero produced RDF triples in complex testbeds can generate 32.65\% of the total number of RDF triples using planning. %Thus, these techniques put in perspective the benefits of planning the execution of mapping assertions, and provide evidence of the need for more efficient data management approaches for scaling up KG creation to larger and heterogeneous data sources.
\end{itemize}
This paper is organized into six additional sections.
Preliminaries and a motivating example are presented in section \ref{sec:preliminaries}, and
related approaches are discussed in section \ref{sec:relatedwork}. 
Section \ref{sec:kgc} presents the problem of KG creation and discusses the proposed execution planning techniques.
The KG creation techniques implemented to execute bushy tree plans are explained in section \ref{sec:arquitecture}. 
Section \ref{sec:experiments} reports on the results of the empirical evaluation.
Lastly, section \ref{sec:conclusions} summarizes lessons learned and outlines future directions.

%%%%%%%%%%%%%%%%%%%%%%%%%%%%%%%%%%%%%%%%
%%             MOTIVATION             %%
%%%%%%%%%%%%%%%%%%%%%%%%%%%%%%%%%%%%%%%%
\section{Preliminaries and Motivation}
\label{sec:preliminaries}
KGs are data structures that model factual statements as entities and their relationships using a graph data model~\cite{gutierrez2021knowledge}. The creation process of a KG $\mathcal{G}$ is defined in terms of a data integration system $DIS_\mathcal{G}=\langle O,S,M \rangle$ where $O$ is a set of classes and properties of a unified ontology, $S$ is a set of data sources, and $M$ corresponds to mapping rules or assertions defining concepts in $O$ as conjunctive queries over sources in $S$. The execution of the $M$ rules over data from sources in $S$ generates the instances of $\mathcal{G}$. Figure \ref{fig:preliminaries} shows mapping assertions represented in RML \cite{DimouSCVMW14}, an extension of R2RML  \cite{r2rml}, the W3C recommendation for mapping rules from data sources in various formats (e.g., CSV and JSON) to RDF. 

\noindent \textbf{Mapping Assertions}.
Mapping rules in $M$ are formalized as Horn clauses \[body(\overline{X}):-head(\overline{Y})\] that follow the Global As View (GAV) approach (Namici et al.~\cite{namici2018comparing}), i.e., $body(\overline{X})$ is a conjunction of predicates over the sources in $S$ and their attributes, and $head(\overline{X})$ is a predicate representing classes and properties in $O$. Variables in $\overline{Y}$ are all in $\overline{X}$, and the rule head may include functions. 
 They correspond to an abstract representation of the triples maps, expressed in mapping languages like R2RML~\cite{das2012r2rml} or RML~\cite{DimouSCVMW14}. There are three types of mapping assertions: concept, role, and attribute.  
  \begin{itemize}
      \item \textbf{Concept Mapping Assertions} are conjunctive rules over the predicate symbols of data sources in $S$ to create the instances of a class $C$ in the ontology $O$.  
      %The identifier of each instance in $C$ can be expressed in terms of functions whose parameters are attributes defined in the body of the rule. 
      Without loss of generality, we assume that the body is composed of only one source. Thus, concept mapping assertions have the form of: \[S_i(\overline{X}):-C(f(y))\] 
      Using the R2RML terminology, a concept mapping assertion corresponds to a \verb|rr:subjectMap| where attributes in the logical source $S_i$, define the subject of the class $C$; $f(.)$ corresponds to a predefined function that enables the concatenation of strings, expressed with the RDF predicate \verb|rr:template|. Figure
      \ref{fig:preliminaries}
      depicts three RML triples maps and their corresponding mapping assertions. These concepts mapping assertions define classes \texttt{C1}, \texttt{C2}, and \texttt{C3}. 
  \item  \textbf{Role Mapping Assertions} enable the definition of object properties or roles. We differentiate three types of role mapping assertions. 
\\ 
\noindent  
\textbf{Single-Source Role Mapping Assertions} define a role $P(.,.)$ in terms of a source's attributes, where $f_1(.)$ and $f_2(.)$ are function symbols: \[S_i(\overline{X}):-P(f_1(y_1),f_2(y_2))\] 
In Figure \ref{fig:preliminaries},
the triples map \texttt{TriplesMap2} defines the property \texttt{ex:p5} as a single-source role assertion.  
The rule \verb|rr:predicateObjectMap| defines  \texttt{ex:p5} object value with \verb|rr:objectMap|; \verb|rr:template| corresponds to a pre-defined function. 
\\ 
\noindent
\textbf{Referenced-Source Role Mapping Assertions} specify the object value of a role $P(.,.)$ over a source $S_i$ that also defines the subject of a referred concept mapping assertion $MA$. 
\[S_i(\overline{X_{i,1}}),S^{MA}_i(\overline{X_{i,2}}): -
P(f_1(y_1),f_2(y_2))\]
\[\textit{MA:} \;\; S_i(\overline{X_{i,2}}):- C_j(f_2(y_2))\]
Using the R2RML terminology, this assertion corresponds to a \verb|rr:RefObjec| \verb|tMap| where the mapping assertion $MA$ is referred using the predicate \verb|rr:par| \verb|entTriplesMap|. Both mapping assertions are defined over the same logical source $S_i$. In Figure \ref{fig:preliminaries}, \texttt{TriplesMap1} defines the property \texttt{ex:p3} as the subject of the triples map \texttt{TriplesMap2}. Both \texttt{TriplesMap1} and \texttt{TriplesMap2} are defined over the same logical source. 
 \begin{figure*}[t!]
    \includegraphics[width=1.0\textwidth]{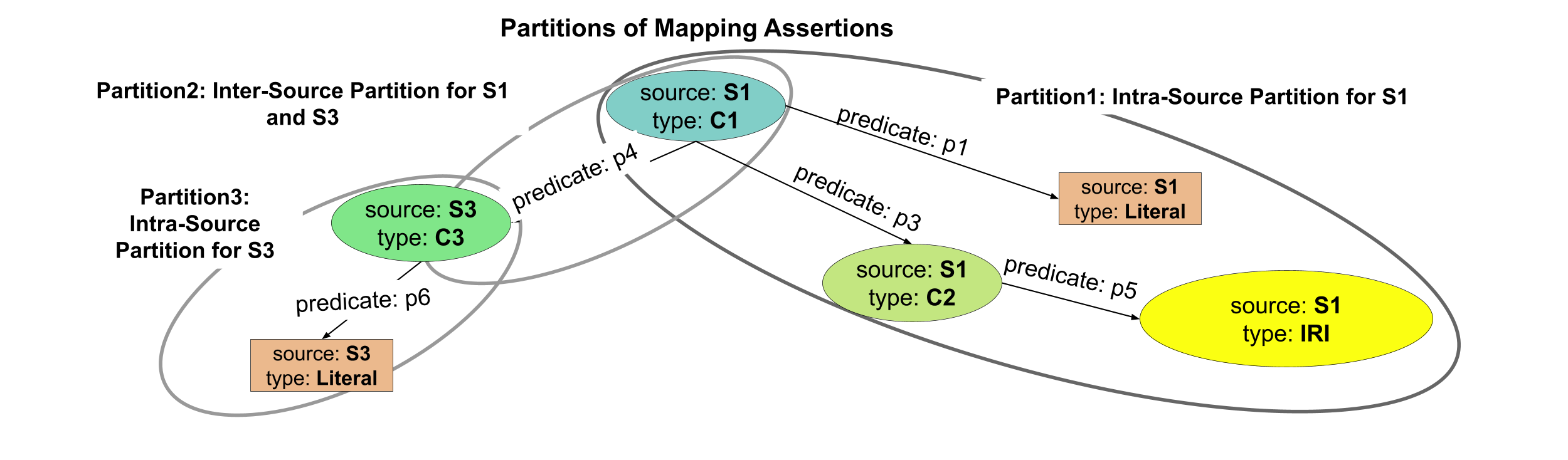}
    \caption{\textbf{Partitioning of Mapping Assertions.} Mapping assertions and partitions.}
    \label{fig:mapping}
\end{figure*}
\noindent
\textbf{Multi-Source Role Mapping Assertions} 
 allow for the definition of a role $P(.,.)$ where the subject and object are defined over different sources, i.e., $S_j$ and $S_i$, respectively. The source $S_j$ is utilized in another assertion $MA$ to define the instances of a class $C_k$. Because the sources, $S_i$ and $S_j$ are different, a join condition is required. The mapping assertion is denoted by the rule, \[S_i(\overline{X_{i,1}}),S^{MA}_j(\overline{X_{i,2}}),\theta(\overline{X_{i,1}},\overline{X_{i,2}}):- \]
\[ P(f_1(y_1),f_2(y_2)) \]
where 
$\theta(\overline{X_{i,1}},\overline{X_{i,2}})$ stands for the join condition. Further, the referred concept mapping assertion $MA$
is as \[\textit{MA:} \;\; S_j(\overline{X_{i,2}}):-C_k(f_2(y_2))\]
Using R2RML terminology, this assertion corresponds to a \verb|rr:RefObjectMap| including \verb|rr:joinCondition|, where $MA$ stands for the triples map referred by the predicate \verb|rr:parentTriplesMap|. In Figure \ref{fig:preliminaries}, \texttt{p4} is defined using a multi-source role mapping assertion that relates \texttt{TriplesMap2} and \texttt{TriplesMap3}. Since both triples maps are defined over two different logical sources, \texttt{S1.csv} and \texttt{S3.csv}, it is required the join condition between the field \texttt{attribute} from \texttt{S1.csv} and the field \texttt{DrugName} to determine which value of the subject of  \texttt{TriplesMap3} will be used as the object value of \texttt{p4}. 
  \item 
\textbf{Attribute Mapping Assertions} express a property $A$ where the subject is defined with a function, and the object value is a literal. The clause following rule represents this assertion,
\[S_i(\overline{X}):-A(f(y_1),y_2)\] 
where, $y_2$ stands for a variable in the list of variables $\overline{X}$ from where the object value of the attribute $A$ is retrieved. The map  \verb|objectMap| inside a \verb|predicateObjectMap| defines the object value as a \verb|rml:reference| or \verb|rr:column|. In Figure \ref{fig:preliminaries}, two attribute mapping assertions specify the attributes \texttt{p1} and \texttt{p6} in \texttt{TriplesMap1} and \texttt{TriplesMap3}, respectively.
  \end{itemize}
  \begin{figure*}[h!]
    \includegraphics[width=1.0\textwidth]{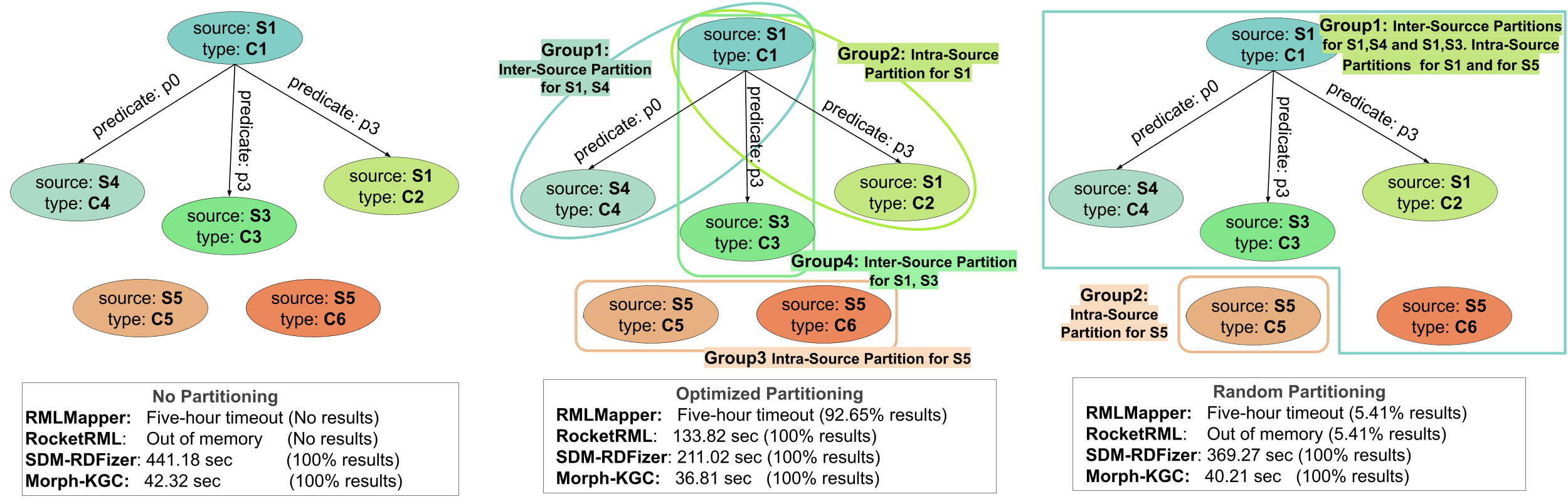}
    \caption{\textbf{Motivating example.} This figure illustrates three possible approaches to executing the motivating example of this work. The left figure presents an approach where the mapping assertions are executed without partitioning. The central figure illustrates the proposed approach, where four partitions are generated. Finally, the right figure presents a random partitioning, only creating two partitions.}
    \label{fig:motivating}
\end{figure*}
\subsection{Partition of Mapping Assertions}
 In a data integration system $DIS_\mathcal{G}=\langle O,S,M \rangle$, the mapping assertions in $M$ can be grouped to create a partition of $M$. We define two types of partitions: \textbf{Intra-source} and \textbf{Inter-source} mapping assertion partitions. 
Given a source $S_k$ in $S$, an \textbf{Intra-source} partition for $S_k$ corresponds to a set of all the mapping assertions that have only the source $S_k$ in the body clause, i.e., it comprises concept, attribute, single-source role, and referenced-source role mapping assertions over $S_k$. An \textbf{Inter-source} groups mapping assertions of two sources $S_i$ and $S_j$ which are related via multi-source role mapping assertions. Figure \ref{fig:mapping} presents three partitions for mapping assertions in the running example. To increase readability, mapping assertions are depicted in a directed graph where directed edges represent predicates defined by mapping assertions (i.e., \texttt{p4}, \texttt{p6}, \texttt{p1}, \texttt{p3}, and \texttt{p5}). A node denotes a logical source and the type of the mapped entity.    
All the assertions defined over $S1$ (resp. $S3$) are grouped together into \texttt{Partition1} (resp. \texttt{Partition3}). Moreover, there is only one assertion between \texttt{S1} and \texttt{S3}, thus, \texttt{Partition2} is an \textbf{inter-source} partition and comprises the multi-source mapping assertion for \texttt{p4} and the concept mapping assertion that defines the class \texttt{C3}. 

\subsection{Motivating Example}
We motivate our work, illustrating the challenges that the execution of mapping assertions brings to the process of KG creation from multiple data sources. Continuous creation and maintenance of KGs demand scalability in terms of required execution time and memory consumption. Figure \ref{fig:motivating} presents three configurations of a set of mapping assertions that define a KG $G1$. The set comprises mapping assertions specifying the properties and attributes of five classes (\texttt{C1},\texttt{C2},\texttt{C3},\texttt{C4}, and \texttt{C5}) over four data sources (\texttt{S1}, \texttt{S3}, \texttt{S4}, and \texttt{S5}). These data sources correspond to the SDM-Genomic-Datasets, each containing one Million records and up to 15 attributes. 
\\\noindent
The configuration \texttt{No Partitioning} depicts all the mapping assertions; they are executed together on four state-of-the-art [R2]RML-compliant engines, RMLMapper~\cite{DBLP:conf/www/DimouNVMW16}, RocketRML~\cite{rocketrml}, SDM-RDFizer~\cite{iglesias2020sdm}, and Morph-KGC~\cite{arenas2022morph}. Executing all the assertions together demands from each engine, data management techniques like the ones implemented by Morph-KGC. These techniques must allow planning both the execution of the mapping assertions and the period to maintain in memory each source. Unfortunately, RMLMapper and RocketRML are not as scalable as Morph-KGC and cannot produce any results. RocketRML ran out of memory, while RMLMapper timed out after five hours. On the contrary,
all the engines exhibit better performance when the assertions are divided into intra- and inter-source partitions and executed in plans generated based on these partitions; the improvement, albeit not so significant as in the other engines, can also be observed in Morph-KGC. First, when four groups of partitions are created (i.e., \texttt{Optimized Partition}), the performance of the four engines is empowered, and three of them can generate 100\% of the results. Each group comprises one intra-source partition of a source $S_j$ and at most one inter-source partition of another source $S_i$ to $S_j$. Moreover, the groups are executed in parallel. Lastly, the execution of the configuration named, \texttt{Random Partition}, indicates that no combination of the intra- and inter-source partitions leads to efficient mapping assertions plans. In this case, \texttt{Group1} includes two inter- and four intra-source partitions, while \texttt{Group2} comprises only one intra-source partition. Although \texttt{Group2} is executed by all the engines, RMLMapper and RocketRML could not produce any result during the execution of \texttt{Group1}, and they could only produce 5.41\% of the total number of RDF triples. This paper addresses the challenges of generating plans of mapping assertions that empower [R2]RML engines and enhance their scalability during KG creation.  
%%%%%%%%%%%%%%%%%%%%%%%%%%%%%%%%%%%%%%%%
%%              Related Work              %%
%%%%%%%%%%%%%%%%%%%%%%%%%%%%%%%%%%%%%%%%
\section{Related Work}
\label{sec:relatedwork}
\subsection{Semantic Data Integration}
A KG creation process relies on semantic data integration frameworks. The seminal work of Lenzerini~\cite{lenzerini2002data} formalizes the fundamentals of data integration systems and the paradigms for mapping heterogeneous data sources into a unified schema. Knoblock et al. propose KARMA~\cite{knoblock2015exploiting} a semi-automatic framework to map structured sources to ontologies and integrate them at the level of schema. There are different novel approaches to integrate generated RDF data, which can be considered as the KG creation post-processing. LDIF is introduced by Schultz et al. ~\cite{bizer2012ldif} which relies on a set of tools including Silk~\cite{isele2013active} and Sieve~\cite{mendes2012sieve} to link identified entities and the data fusion tasks, respectively. MINTE~\cite{collarana2017minte}, a semantic RDF data integration technique, is proposed by Collarana et al., relying on the metadata about the classes and properties to integrate semantically equivalent entities, while, Benbernou et al. define an approach for RDF data integration~\cite{benbernou2017semantic}. In the case of big data, post-processing integration is only affordable if the main KG creation framework is efficient. In other words, if the generation of RDF triples is expensive, any semantically duplicated RDF triples must be integrated prior to KG creation to improve scalability and efficiency. 
\subsection{Mapping Languages and KG Creation Frameworks}
A KG can be generated by semantifying and integrating heterogeneous data into an RDF data model; different tools and approaches can be applied for this purpose. In order to provide a flexible and transparent transformation, declarative mapping languages are proposed to map the data into the concepts of the unified schema or the ontology and transfer them into RDF. R2RML~\cite{r2rml} recommended by the World Wide Web Consortium (W3C) and RDF Mapping Languages (RML), the extension of R2RML, are two popular exemplar declarative mapping languages. Accordingly, several methods and tools are proposed for transforming data into RDF using R2RML and RML mapping rules such as RMLMapper~\cite{dimou_ldow_2016}, SDM-RDFizer~\cite{iglesias2020sdm}, RocketRML~\cite{rocketrml}, and CARML~\cite{carml}. Priyatna et al.~\cite{DBLP:conf/www/PriyatnaCS14} introduce an extension of an existing SPARQL to SQL query rewriting algorithm, applying R2RML mapping rules. As a different approach, Lefracios et al.~\cite{lefranccois2017sparql} propose an extension of SPARQL named SPARQL-Generate to generate RDF. In order to scale up the process of transforming data into RDF and creation of KG for large or complex data integration systems, different optimization frameworks are proposed, some of which can be applied along with mentioned tools. For instance, Szekely et al. propose the DIG system~\cite{szekely2015building}, Jozashoori and Vidal define MapSDI~\cite{jozashoori2019mapsdi}, while Gawriljuk et al.~\cite{gawriljuk2016scalable} present a scalable framework for incremental KG creation. Morph-KGC~\cite{arenas2022morph} proposes an approach to partition R2RML and RML mapping assertions so that generated partitions can be executed in parallel. Morph-KGC relies on partitioning the mapping assertions into groups that generate disjoint sets of RDF triples. Nevertheless, based on this partitioning strategy, RDF triples with a \emph{join dependency}, i.e., the subject of one RDF triple is the object of another, are partitioned into independent groups. Therefore, the same join RDF resource is generated redundantly by each disjoint partition to ensure the completeness and correctness of the result RDF triples. Nevertheless, an efficient partitioning strategy requires considering all mapping assertions including those that generate RDF triple sets with join dependency as a whole, to ensure that the result partitions are optimized.
Therefore, despite the significance of all mentioned contributions and improvements, none of the mentioned approaches addresses the problem of scheduling the optimized execution of mapping assertion partitions, specifically considering different impacting factors, e.g., mapping assertions types, connection between mapping assertions, and common properties among them. Additionally, the mentioned approaches are specific for an engine, i.e., they are not necessarily adaptable to generic KG creation pipelines. We tackle the mentioned existing limitations, introducing an engine-agnostic execution technique relying on efficient partitioning and scheduling strategies. The proposed execution planner decides on the optimized execution plan based on the types of mapping assertions, the connection between the mapping assertions, and the redundancy of the predicates in mapping assertions. Any [R2]RML-compliant engine can adopt our proposed optimization approach, as shown in the next sections.
\subsection{KG Creation from Textual Data}
\label{subsec:textual}
\noindent Integrating semi/unstructured data, e.g., texts, and constructing KGs from such data requires a semantic layer to describe the data and further data manipulation/transformation steps such as data cleaning, Named-Entity Recognition (NER), and Entity Linking (EL). Chessa et al. introduce ~\cite{chessaenriching} a methodology to add a semantic layer to a data lake and create a KG. Barroca et al.~\cite{barroca2022enriching} extract metadata from textual descriptions and link them to entities in KGs utilizing NER and EL techniques, while Chu et al. propose a method to address the challenge of entity relations extraction~\cite{chu2021knowfi}. Additionally, data manipulation/transformations can also be defined in terms of functions as part of declarative mapping assertions applying the available extensions including RML+FnO~\cite{de2016ontology}, R2RML-F~\cite{debruyne2016r2rml}, FunUL~\cite{junior2016funul}, and D-REPR~\cite{vu2019d}. In this regard, EABlock~\cite{jozashoori2022eablock} provides a library of FnO functions that perform entity alignment on the input entity value, relying on an engine implementing the tasks of NER and EL. Considering the importance of efficiency in KG creation, FunMap~\cite{jozashoori2020funmap} proposes efficient executions of FnO functions. The techniques proposed in this paper are illustrated and evaluated in mapping assertions over structured data. Nevertheless, they can be applied with approaches like FunMap to speed up the KG creation from unstructured data. 
\begin{table*}[h!]
\caption{Notation Summary}\label{tab:notations}
\scriptsize
\centering
\begin{tabular}{|m{3.5cm}|m{11.5cm}|}
\hline
\rowcolor{blue!30}
\textbf{Notation} & \textbf{Explanation} \\ \hline
 $DIS_\mathcal{G}=\langle O,S,M \rangle$ &  Data Integration System, where $O$ is a unified ontology, $S$ is a set of data sources, and $M$ corresponds to mapping assertions defining concepts in $O$ over sources in $S$. The execution of rules in $M$ over data sources in $S$ generates the knowledge graph $\mathcal{G}$.\\ \hline
 \textit{body}($\overline{X}$):-\textit{head}($\overline{Y}$) &  Mapping Assertion in $M$ defined as Horn clauses; $body(\overline{X})$ is a conjunction of predicates over the sources in $S$ and their attributes, and $head(\overline{X})$ is a predicate representing classes and properties in $O$.\\ \hline
$S_i(\overline{X})$ & Predicate symbol for data source in $S$ with arguments $\overline{X}$.\\ \hline
$C(f(y))$ & Predicate symbol for class in $O$; $f(y)$ functional symbol with arguments $y$.\\ \hline
$P$($f_1$($y_1$),$f_2$($y_2$)) & Role predicate in $O$; $f_1$($y_1$) and $f_2$($y_2$) are functional symbols.\\ \hline
$S^{MR}_i$($\overline{X_{i,2}})$ & Predicate symbol representing data source in the body of mapping assertion $MR$\\ \hline
$\theta(\overline{X_{i,1}},\overline{X_{i,2}})$ & Join condition between the attributes of predicate symbols\\ \hline
$A(f(y_1),y_2)$ & Predicate symbol for a data property; $f(y_1)$ functional symbol.\\ \hline
$GP_{M}$ & Set of sets of mapping assertions in $M$\\ \hline
$\overline{GP}_{M}$ & Plan over groups of mapping assertions in $GP_{M}$.\\ \hline
$BT$ & Bushy Tree plan of groups of mapping assertions.\\ \hline
 $OP$ & Binary operator in a bushy tree. \\ \hline
\texttt{DR} & Union with Duplicate Removal\\ \hline
\texttt{NDR} & No-Duplicate Removal Union\\ \hline
$fu(.,.)$ & Utility function for quantifying a bushy tree plan performance\\ \hline
$\mathcal{B}^{GP_{M}}$ & Set of the bushy trees over $GP_{M}$\\ \hline
$\mathcal{SS}$ & Power set of $SS$\\ \hline
$\delta(G_i)$ & Execution cost of group of mapping assertions $G_i$\\ \hline
$ma_j$ & Mapping assertion on source $S_j$\\ \hline
\end{tabular}
\end{table*}
\subsection{Benchmarking KG Creation} 
Namici et al.~\cite{namici2018comparing} compare two state-of-the-art engines in Ontology-Based Data Access by formalizing the two systems, considering W3C-compliant settings. In addition to the theoretical efforts, empirical evaluations such as the study by Chaves et al.~\cite{Chaves-FragaEIC19} are conducted to define the parameters affecting KG creation. Accordingly, benchmarks that consider the impacting parameters~\cite{Chaves-FragaEIC19} are required to assess and compare the performance of different KG pipelines. 

One of the proposed benchmarks to evaluate different Ontology-Based Data Integration or KG creation frameworks is GTFS-Madrid-Bench~\cite{chaves2020gtfs}; this benchmark provides a set of heterogeneous data and mappings. Although GTFS-Madrid-Bench promises to ensure diversity, this benchmark lacks the requirements for studying all the impacting parameters reported in~\cite{Chaves-FragaEIC19}. 
For instance, to evaluate the impact of data volume on different KG creation approaches, it is essential to have an equal growth of the volume in all the datasets involved in the KG; however, this requirement is not met by GTFS-Madrid-Bench. Furthermore, the deficiency of required testbeds to study parameters such as join selectivity, star-join, data duplicates, and duplicated predicates in mappings is another limitation of GTFS-Madrid-Bench. 
Therefore, to ensure the fairness and comprehensiveness of our experimental study, in addition to GTFS-Madrid-Bench, we also consider and extend SDM-Genomic-Datasets~\cite{sdm-genomic} to include other impacting parameters that affect KG creation scalability (e.g., complexity of mapping assertions and percentage of duplicates).
\section{Scaling KG Creation Up}
\label{sec:kgc}
\begin{figure*}[h!]
    \centering
    \begin{subfigure}{.45\linewidth}
    \centering
        \includegraphics[width=1.0\linewidth]{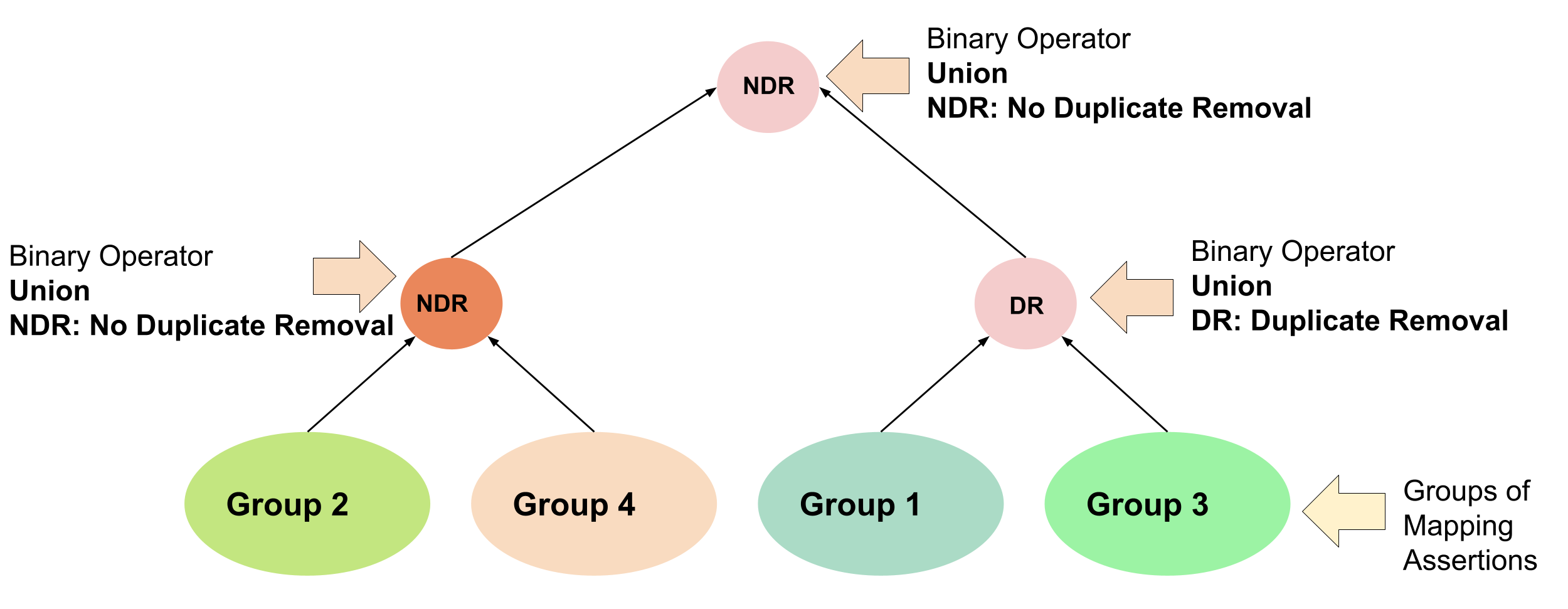}
   %\vspace{-0.25cm}
    \caption{Bushy Tree Plan}
  \label{fig:bushyTreeA}
    \end{subfigure} 
    %\hspace{-0.6cm}
     \begin{subfigure}{.45\linewidth}
      \centering
        \includegraphics[width=1.0\linewidth]{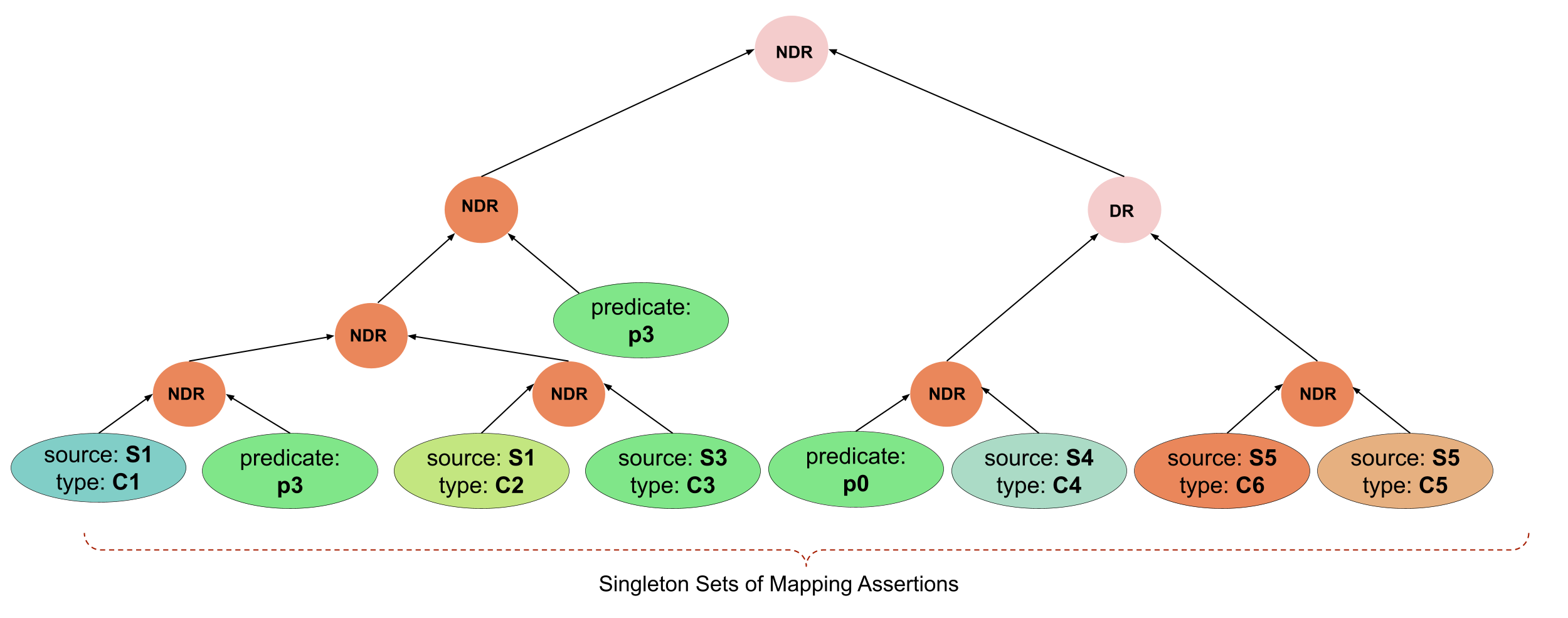}
        %\vspace{-0.25cm}
         \caption{Simple Bushy Tree Plan}
        \label{fig:simpleTree}
\end{subfigure} 
 \caption{Bushy Tree Plans of Mapping Assertions. a) Tree Plan whose leaves are intra- and inter-source groups of mapping assertions. b)Tree Plan whose leaves are singleton sets of mapping assertions. The simple bushy tree plan in (b) requires the execution of more union operators and loading the data sources multiple times than the execution of the bushy tree plan in (a).}
    \label{fig:trees}
\end{figure*}
This section formalizes the problem tackled in this paper and presents the proposed solution; the notation used in the formalization is summarized in \autoref{tab:notations}. 
\\\noindent
The process of creating a KG $\mathcal{G}$ is defined as a data integration system $DIS_\mathcal{G}=\langle O,S,M \rangle$, where mappings in $M$ correspond to assertions defined in [R2]RML. As observed in Figure \ref{fig:motivating}, 
the order and grouping of the mapping assertions impact the execution time of the engines, which is crucial to enable the generation of results in real-world scenarios.
The aim is to generate $GP_{M}$, a set of sets of mapping assertions in $M$ (inter- and intra-source), such as the union of all the sets in $GP_{M}$ is equal to $M$, and the pair-wise intersection of the sets in $GP_{M}$ is empty.
That is, $GP_{M}$ is a partition of $M$.
Moreover, since the order in which the groups in $GP_{M}$ may also impact, we define a plan $\overline{GP}_{M}$ over the groups in $GP_{M}$, as a bushy tree plan of the groups in $GP_{M}$, where each internal node represents the union operator that merges the RDF triples produced during the execution of each group in $GP_{M}$.
Lastly, since results produced during the execution of the $GP_{M}$ groups may overlap, and duplicate removal may be required at different steps of the execution of $\overline{GP}_{M}$.
Thus, each node is annotated with the union operator, which merges the inputs and produces the results.

\begin{figure*}[t!]
    \centering
 \includegraphics[width=1.0\textwidth]{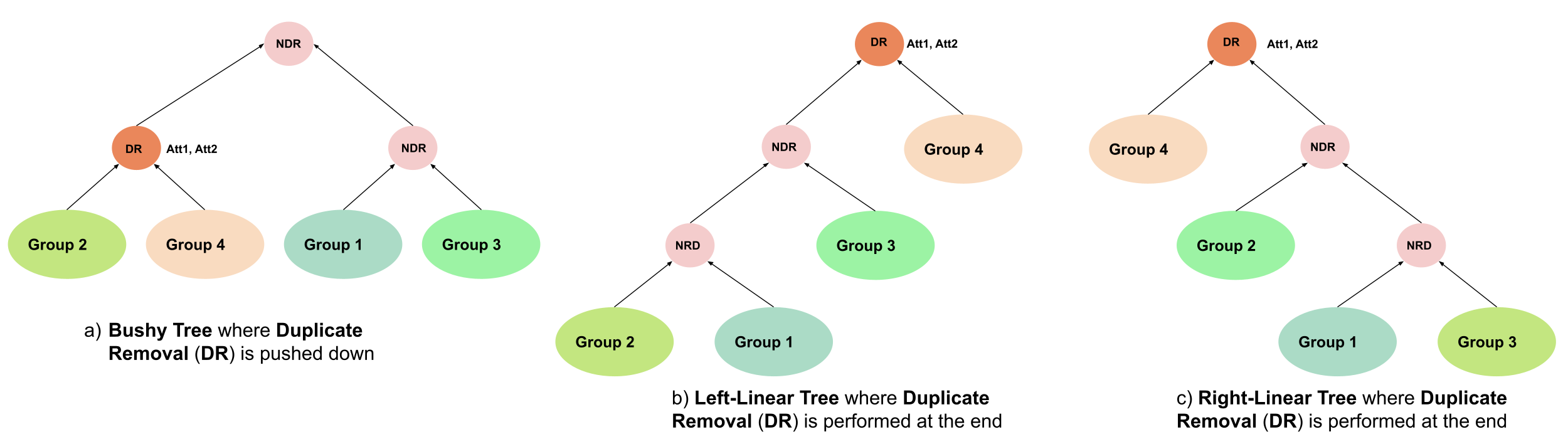}
  \caption{\textbf{Running example.} Execution Trees for the groups in the Optimized Partition in Figure \ref{fig:motivating}. Bushy tree in a) performs an eager duplicate removal, while the evaluation of the duplicate removal is lazy in the tree plans in Figures b) and c).}
    \label{fig:bushyTree}
\end{figure*} 

A \emph{bushy tree} is a data structure corresponding to a binary tree. As in regular trees, nodes with no children are called leaves, and the root node does not have any parent node. Additionally, in bushy trees, non-leaf nodes have exactly two children, and all the nodes, except the root, have one single parent node \cite{DBLP:conf/pods/ScheufeleM97}. 
A \emph{plan} $BT$ over groups of mapping assertions is a bushy tree; it is inductively defined as follows: 
 \\\noindent \textbf{Base Case.} Let $BT$ be a 
 group of mapping assertions. $BT$ is a bushy tree plan which corresponds to a leaf.
  \\\noindent \textbf{Inductive Case}.
Let $BT1$ and $BT2$ be bushy tree plans over groups of mapping assertions. Let $OP$ be a binary set operator (e.g., union),  then the following is a bushy tree plan over groups of mapping assertions:
\begin{center}
\begin{forest}
[OP
        [BT1]      
        [BT2]
]  
\end{forest}
\end{center}

A plan $\overline{GP}_{M}$\footnote{We use $BT$ and $\overline{GP}_{M}$ to denote bushy tree plans over mapping assertions. $BT$ represents a generic plan, while $\overline{GP}_{M}$ is specifically for the groups of mapping assertions in $GP_{M}$. } is a bushy tree plan where the groups of mapping assertions in $GP_{M}$ are the $\overline{GP}_{M}$ leaves. The binary operators in $\overline{GP}_{M}$ correspond to the union of sets. They can be \texttt{DR} union with duplicate removal, or \texttt{NDR} non-duplicate removal union. 

Additionally, the leaves of a bushy tree plan can correspond to intra- or inter-source partitions in $GP_{M}$. However, the leaves of a bushy tree can also comprise one mapping assertion; we call these plans, \emph{simple bushy tree} plans. Figure \ref{fig:trees} depicts two bushy trees over the mapping assertions of the motivating example presented in Figure \ref{fig:motivating}. The bushy tree plan in \autoref{fig:bushyTreeA} comprises four intra- and inter-source groups of mapping assertions. Contrary, the leaves in the bushy tree in \autoref{fig:simpleTree} correspond to singleton sets composed of one mapping assertion. The execution of the simple bushy tree plan requires the evaluation of more union operators and loading, several times, data sources \textbf{S1} and \textbf{S5} in main memory. 

An \emph{optimal bushy tree plan} is a bushy tree plan whose evaluation is duplicate free, and its execution cost is minimal. 
Moreover, the evaluation of the duplicate removal operators can be \emph{eager} or \emph{lazy}. Intuitively, an eager evaluation of a duplicate removal union is performed in a bushy tree as soon as the duplicates are produced. Thus, 
the execution of the operator $OP$ in a bushy tree $BT$ that unions subtrees $BT_1$ and $BT_2$ 
is an \emph{eager duplicate removal}, if the execution of $BT_1$ and $BT_2$ is duplicate-free, but the intersection between $BT_1$ and $BT_2$ is not empty. On the other hand, a \emph{lazy evaluation} of the duplicate removal receives input collections with duplicates and removes duplicates from the union of the two inputs. Thus, in a bushy tree plan $BT$ with lazy evaluation, $BT_1$ and $BT_2$ are not duplicate free because the duplicate removal operator has been postponed. 

There are $\frac{(2n-2)!}{(n-1)!}$ bushy trees $\overline{GP}_{M}$, where $n$ is the cardinality of $GP_{M}$ \cite{DBLP:conf/pods/ScheufeleM97}. 
Figure \ref{fig:bushyTree} depicts three bushy trees for the groups of the \texttt{Optimized Partition} presented in the motivating example depicted in Figure \ref{fig:motivating}. 
Figure \ref{fig:bushyTree}a) illustrates a bushy tree where \texttt{DR} is pushed down into the tree, scheduling, thus, this operation in a smaller RDF triple set.
Note that \texttt{Group 2} and \texttt{Group 4} comprise mapping assertions that define instances of the class \texttt{C1} and the property \texttt{p3}. As a result, the merge of RDF triples produced during the execution of these groups may contain duplicates that need to be eliminated, and the duplicate removal operator  \texttt{DR} is required. Since the duplicate removal is executed as soon as the duplicates are generated, the execution is eager. 
Contrary, mapping assertions in \texttt{Group 1} and \texttt{Group 3} do not commonly define any class or predicate; thus, \texttt{NDR} is the union operator between them.  Figures \ref{fig:bushyTree}b) and c) correspond to
left- and right-linear plans. Additionally, duplicated removal is performed over the whole set of RDF triples, i.e., this is a lazy evaluation of the duplicate removal. The execution of these plans may require more memory and execution time in comparison to the execution of the bushy plan in Figure \ref{fig:bushyTree}a). 

A utility or cost function can quantify the performance of a bushy tree plan. The function $fu(.,.)$ measures execution time or memory consumption; it is a \emph{lower-is-better} function, i.e., the lower the execution cost, the better the plan performance. Let $\mathcal{B}^{GP_{M}}$ be the set of the bushy trees over $GP_{M}$, and let $\mathcal{SS}$ be the power set of $S$:   \[fu:\mathcal{B}^{GP_{M}} \times \mathcal{SS} \rightarrow \mathbf{R}\]
$fu(.,.)$ is inductively defined on the structure of a bushy tree plan $BT$ as follows. 

 \noindent \textbf{Base Case.} Let $G_i$ be a 
 group of mapping assertions on data sources in $S$ and the assertions in $M$, such that $G_i$ is a leaf of $BT$
 \[fu(G_i,S)=\delta(G_i)\] 
 where, $\delta(G_i)$ represents the execution cost of $G_i$. 
 In our experiments, $\delta(G_i)$ corresponds to the elapsed time required to execute the mapping assertions in $G_i$ and store the generated RDF triples in secondary memory. Also, $\delta(G_i)$ can quantify memory consumption, and be defined as the amount of main memory consumed during the execution of $G_i$. 
 Alternatively, Iglesias et al. \cite{iglesias2020sdm} presents an abstract cost function defined in terms of the number of comparisons and insertions in main-memory data structures required for executing $G_i$. They represent possible implementations of $\delta(G_i)$.
 \\\noindent \textbf{Inductive Case}.
Let $BT$ be a bushy tree plan composed by the union operator $OP$ that merges the results of executing the bushy tree plans $BT_1$ and $BT_2$:
\begin{equation}
\begin{aligned}
fu(BT,S)={} & fu(BT_1,S) + fu(BT_2,S) +  \\
         &\phi(OP,BT_1,BT_2)
\end{aligned}
\end{equation}

\noindent $\phi(OP,BT_1,BT_2)$ corresponds to the cost of executing $OP$ over the RDF triples produced by the execution of $BT_1$ and $BT_2$. If $OP$ is the duplicate removal operator \texttt{DR}, the time complexity is $O(N \log N)$, where $N$ is the sum of the size of the RDF triples produced by the execution of $BT_1$ and $BT_2$. Otherwise, $\phi(OP,BT_1,BT_2)$ is $O(N)$ \cite{TeuholaW91}.
\subsection{Problem Statement} 
Let $DIS_\mathcal{G}=\langle O,S,M \rangle$, $GP_{M}$, and $\mathcal{B}^{GP_{M}}$ be, respectively, a data integration system,    a partition of $M$, and the set of all the bushy trees $\overline{GP}_{M}$ over $GP_{M}$. Consider a utility function, $fu(\overline{GP}_{M},S)$, that computes the cost of executing $\overline{GP}_{M}$ over sources in $S$.   

The problem of \emph{planning KG creation} corresponds to finding the bushy tree $\overline{GP}_{M}$ whose execution in $S$ minimizes $fu(\overline{GP}_{M},S)$ and  creates the duplicate-free RDF triples in $\mathcal{G}$. $\overline{GP}_{M}$ should satisfy the following conditions:
\begin{itemize}
\item The execution of $\overline{GP}_{M}$ over the sources in $S$ is correct and complete, i.e., the execution of the mappings in $M$ following the plan $\overline{GP}_{M}$ produces all the RDF triples in $\mathcal{G}$. 
\item The value of $fu(\overline{GP}_{M},S)$ is minimal, i.e., if $\mathcal{B}^{GP_{M}}$ is the set of the bushy tree plans over $GP_{M}$, then $\overline{GP}_{M}$ is the plan in  $\mathcal{B}^{GP_{M}}$ that minimizes $fu(.,.)$.
\begin{equation} \label{eq:1}
B= \operatorname*{arg\,min}\limits_{\overline{GP}_{M} \in \mathcal{B}^{GP_{M}}}  \text{\textit{fu}}(\overline{GP}_{M},S)
\end{equation}
\end{itemize}

\noindent\textbf{Complexity}.
The problem of constructing a bushy tree plan $\overline{GP}_{M}$ is NP-Hard \cite{DBLP:conf/pods/ScheufeleM97}. 

\subsection{Optimality assumptions}
Finding an optimal tree plan can be done using a cost- or heuristic-based approach. The latter optimization approach requires the definition of a cost model that estimates the cost of each bushy tree plan in $\mathcal{B}^{GP_{M}}$. Alternatively, a heuristic-based method is guided by optimality principles and a set of rules to identify low-cost execution plans. In this work, we present a heuristic-based method to solve the problem of \emph{planning KG creation}. Our proposed method relies on the following optimality principles: 
\begin{itemize}
\item \textbf{P1-Optimality of Intra-Source Partitions.} Let $BT_i$ be a bushy tree with only one leaf, which corresponds to an intra-source partition $G_k$ over a source $S_i$. Let $BT'_i$ be a simple bushy tree for the mapping assertions in $G_k$. The principle of optimality \textbf{P1} assumes that $fu(BT_i,\{S_i\})\leq fu(BT'_i,\{S_i\})$.

\item \textbf{P2-Optimality of Inter-Source Partitions} Let $BT_{i,j}$ be a bushy tree with only one leaf, which corresponds to an inter-source partition $G_{i,j}$ over two sources $S_i$ and $S_j$. Let $BT'_{i,j}$ be a simple bushy tree for the mapping assertions in $G_{i,j}$. The principle of optimality \textbf{P2} assumes that $fu(BT_{i,j},\{S_i,S_j\})$ $\leq fu(BT'_{i,j},\{S_i,S_j\})$.
\item \textbf{P3-Optimality of Bushy Trees.} 
Let $BT$ be a bushy tree over the data sources $S$. $BT$ is of the form
\begin{center}
\begin{forest}
[OP
        [$BT_1$]      
        [$BT_2$]
]  
\end{forest}
\end{center}
bushy plans $BT_1$ and $BT_2$ are optimal, i.e., $fu(BT_1,S)$ and $fu(BT_2,S)$ are minimal and the evaluations of $BT_1$ and $BT_2$ are duplicate free. The principle \textbf{P3} assumes that $BT$ is optimal. 
 
\item \textbf{P4-Optimality of Duplicate Removal} Let $\overline{GP}_{M}$  be a bushy tree plan of mapping assertions in $GP_{M}$ and over data sources in $S$.
Let $\overline{GP'}_{M}$ be an eager duplicate-removal plan of $\overline{GP}_{M}$. Let $\overline{GP''}_{M}$  be lazy duplicate-removal plan of $\overline{GP}_{M}$. The principle \textbf{P4} assumes that 
$fu(\overline{GP'}_{M},S)\leq fu(\overline{GP''}_{M},S)$. 
\end{itemize} 

\begin{figure*}[t!]
    \centering
 \includegraphics[width=1.0\textwidth]{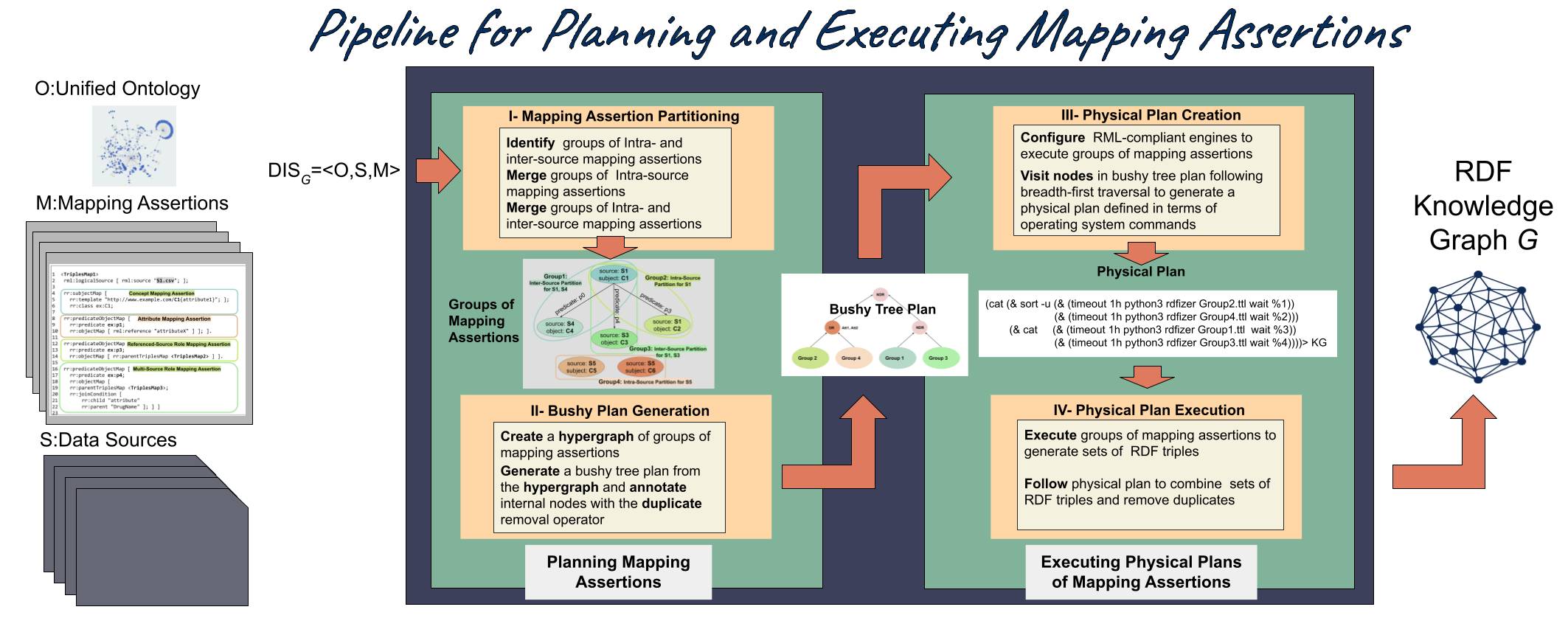}
  \caption{\textbf{Pipeline Steps.} The pipeline receives a data integration system $DIS_\mathcal{G}=\langle O,S,M \rangle$ and outputs a knowledge graph $G$ that corresponds to the execution of the mapping assertions in $M$ over the instances of the data sources in $S$. During the planning phase, $M$ is partitioned into a set of intra- and inter-source mapping assertions and the partition groups are scheduled into a bushy tree plan; the tree internal nodes are annotated with the union operator and duplicate removal is scheduled to be executed as soon as possible. The bushy tree is translated into a physical plan during Executing Physical Plans of Mapping Assertions; this plan states the commands at the operating system that need to be executed for KG creation.   }
\label{fig:arquitecture}
\end{figure*}

Principles \textbf{P1} and \textbf{P2} can be easily demonstrated because a simple bushy tree plan will require uploading in memory several times the same source, increasing, thus, the execution time of evaluating the plan and the amount of consumed memory. 

\noindent Similarly, the proof of principle \textbf{P4} is supported by the cost of the duplicate removal operator, which depends on the size of the multiset from where duplicates will be removed. The cardinality of the result of executing a bushy tree plan $BT$ grows monotonically in terms to the cardinality of its sub-plans $BT_1$ and $BT_2$. Thus, the cost of the eager execution of duplicate removal operators is lower or equal to the lazy execution of the operators. 
Lastly, the principle \textbf{P3} can be ensured based on the optimality of the input sub-plans $BT_1$ and $BT_2$.

Despite the validity of these optimality principles, the outcomes of an optimization method guided by these assumptions can produce plans that are not optimal.

\autoref{theo:optimality} demonstrates the characteristics of a data integration system that ensure the optimality of applying \textbf{P1}-\textbf{P4}. The proof is in \autoref{sec:optimality}. 

\begin{theorem}
\label{theo:optimality}
Let $DIS_\mathcal{G}=\langle O,S,M \rangle$ be a data integration system such that assertions in $M$ meet the following conditions:
\begin{itemize}
\item A concept mapping assertion $ma_j$ on source $S_j$ is referred from any number of multi-source role mapping assertions $ma_i$, but these assertions are all from one source $S_i$. 
\item A property $p$ from $O$ is defined, at most, on one mapping assertion $ma_i$. 
\end{itemize}
Let $BT$ be a bushy tree plan over mapping assertions in $M$ and data sources in $S$; $BT$ generates $G$ and respects the optimality principles \textbf{P1}-\textbf{P4}. Then, $BT$ is optimal, i.e., there is no other equivalent bushy tree plan $BT'$ such as $fu(BT',S) < fu(BT,S)$.
\end{theorem}

\subsection{Proposed Solution} 
We propose a heuristic-based approach to generate a bushy tree $\overline{GP}_{M}$ that corresponds to a solution to the problem of \emph{planning KG creation}. This approach relies on optimality assumptions \textbf{P1}-\textbf{P4}. Thus, the execution of intra- and inter-source groups of mapping assertions independently induces source-based scheduling of the execution of the mapping assertions. At most, two sources are traversed during the evaluation of a group, and less memory is required to keep intermediate results. Lastly, the duplicate removal operators are pushed down into the bushy tree following an eager execution of duplicate removal. As a result, the union operators are scheduled over small sets of RDF triples, and the effect of merging multisets of RDF triples is mitigated. Then,
$\overline{GP}_{M}$ is translated into a physical plan defined in terms of operating system commands. It schedules the execution of each group of mapping assertions and union operators according to $\overline{GP}_{M}$.

\section{The Pipeline for Planning and Executing Mapping Assertions}
\label{sec:arquitecture}
This section describes the techniques that implement the proposed solution reported in the previous section.
Figure \ref{fig:arquitecture} depicts the components of the pipeline for planning and executing a bushy tree $\overline{GP}_{M}$ for the creation of the KG $\mathcal{G}$  defined as a data integration system $DIS_\mathcal{G}=\langle O,S,M \rangle$. 
The pipeline comprises, first, the phase of planning where the bushy tree is created, and then, the execution phase, where $\overline{GP}_{M}$ is translated into a physical plan and executed over a particular [R2]RML-compliant engine.

\subsection{Planning Mapping Assertions}
This step comprises the components of mapping assertion partitioning and bushy plan generation. The algorithm receives a data integration system $DIS_\mathcal{G}=\langle O,S,M \rangle$ and partitions $M$ into groups of intra- and inter-source mapping assertions. Then, they are heuristically combined into a bushy tree plan. These components are guided by the optimality principles \textbf{P1}-\textbf{P4}.

\subsubsection{Mapping Assertion Partitioning} 
The algorithm \textit{Grouping Mapping Assertions} receives as input the set of mapping assertions $M$ and initializes $GP_{M}$ with the intra- and inter-source mapping assertion partitions of $M$. Then, the algorithm \emph{greedily} decides to combine two groups $g_i$ and $g_j$ in $GP_{M}$ into a group $g_{i,j}$ whenever any of the following conditions is satisfied: 
\begin{itemize}
\item \emph{Merging Intra-Source Partitions}. 
This step is guided by the optimality principle \textbf{P1}. Suppose $g_i$ and $g_j$ only comprise intra-source mapping assertion partitions of sources $S'$ (i.e., $S'$ $\subseteq$ $S$). Additionally, there are no sources $S_i$ and $S_j$ in $S'$ such that there exists in $GP_{M}$ an inter-source assertion mapping partition for $S_i$ and $S_j$. Then, groups $g_i$ and $g_j$ can be merged into the group $g_{i,j}$ in $GP_{M}$; $g_{i,j}$ comprises intra-source assertion mapping partitions in $g_i$ and $g_j$. 
\item \emph{Merging Inter- and Intra-Source Partitions.}  This step is guided by the optimality principle \textbf{P2}. Suppose the group $g_i$ comprises an inter-source mapping partition for $S_i$ and $S_j$, where $S_j$ is the referenced source (i.e., logical source of the parent triples map). Additionally, the group $g_j$ only includes the intra-source mapping assertion of $S_j$. Thus, $g_i$ and  $g_j$ can be merged into the group $g_{i,j}$ in $GP_{M}$. The group $g_{i,j}$ only includes intra-source assertion mapping partitions of $S_j$ and the inter-source partition for $S_i$ and $S_j$. In case $S_j$ is the referenced source of various inter-source mapping partitions, the intra-source mapping assertion partition of $S_j$ is only combined with one inter-source 
partition. The selection is done randomly. The selected combination of the intra- and inter-source mapping partitions may be more expensive than other options.  As a result, this decision may negatively impact the performance of a bushy tree plan.
\end{itemize}
\begin{figure*}[t!]
    \centering
 \includegraphics[width=1.0\textwidth]{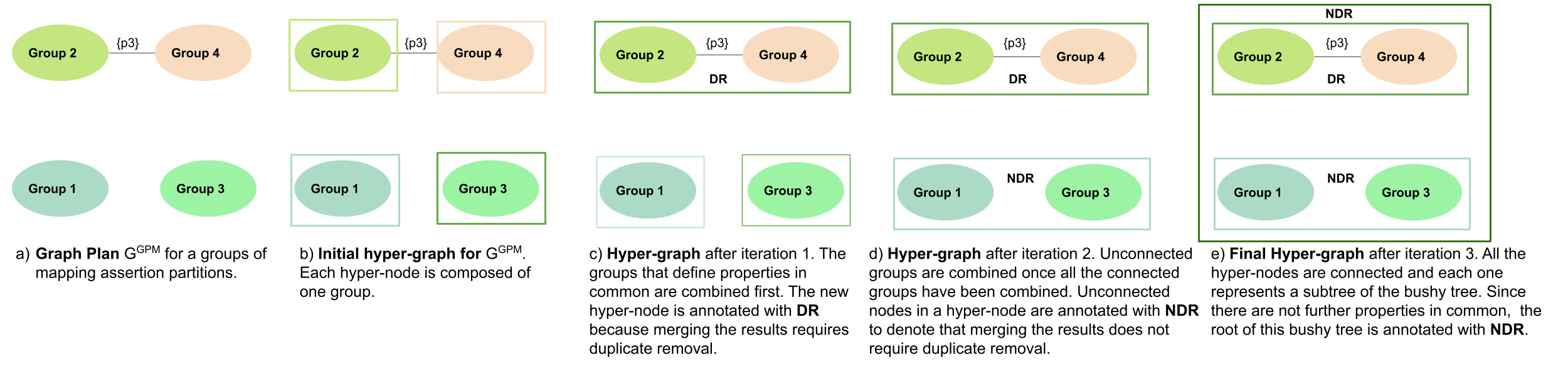}
  \caption{\textbf{Running example.} The Graph Plan for Optimized Partition illustrated in Figure \ref{fig:motivating} is applied, and then the Intermediate Hyper-graphs are generated by the Algorithm Generating a Bushy Tree of Mapping Assertions presented in Figure \ref{fig:bushyTree}.}
\label{fig:exAlgo1}
\end{figure*}
The algorithm iterates until a fixed-point is reached over $GP_{M}$, i.e., an iteration of the algorithm where all the pairs of groups $g_i$ and $g_j$ are revised, and no new group $g_{i,j}$ can replace them in $GP_{M}$.
\subsubsection{Generating a Bushy Tree} 
A bushy tree $\overline{GP}_{M}$ for the groups $GP_{M}$ of mapping assertion partitions is generated following a greedy heuristic-based algorithm; it is guided by the optimality principle \textbf{P3} and assumes that sub-plans produced so far, are optimal. 
Also, the algorithm follows the optimality principle \textbf{P4} and combines first groups of partitions whose union requires duplicate removal. 
\\\noindent
A sketch of the algorithm is outlined in Algorithm \ref{alg:bushytree}. It traverses the set $\mathcal{B}^{GP_{M}}$ in iterations and outputs a bushy tree $B$ where duplicate removal nodes are pushed down. The algorithm receives a graph plan $G^{GP_{M}}$ of the groups in $GP_{M}$ and resorts to a hyper-graph to represent the bushy tree plan $\overline{GP}_{M}$. 
\begin{algorithm}[h!]
\caption{Generating a Hyper-graph of Mapping Assertions.}
\label{alg:bushytree}
\scriptsize
\begin{algorithmic}
\Require $ \text{Plan Graph } G^{GP_{M}}=(V,E,\phi) $
\Ensure $\text{Hyper-graph of Mapping Assertions}$ $OL$
\State $OL \gets empty$
\For{$group \in V$}
\State $OL \gets OL.append(group)$
\EndFor
\State $OL \gets sortByDegree\&NumberSharedProperties(OL)$
\State $FixedPoint \gets FALSE$
\While{$not(FixedPoint)$}
\State $FixedPoint \gets TRUE$
\State $HN  \gets getFirst(OL)$
\State $BestNeighbor \gets getFirstNeighbor(HN)$
\If{$BestNeighbor$ is not NULL}
   \If{$BestNeighbor$ and $HN$ share properties} 
    \State $NewHN \gets merge(HN,BestNeighbor,DR)$
   \Else 
    \State $NewHN \gets merge(HN,BestNeighbor,NDR)$
    \EndIf
    \State $OL.remove(HN)$  
    \State $OL.remove(BestNeighbor)$
    \State $OL.append(NewHN)$
    \State $FixedPoint \gets FALSE$
\EndIf
\EndWhile \\
\Return $OL$
\end{algorithmic}
\end{algorithm}
\noindent A graph plan $G^{GP_{M}}$ is an undirected labelled graph $G^{GP_{M}}=(V,E,\phi)$:
\begin{itemize}
\item The groups in $GP_{M}$ are the nodes in $V$.  
\item There is an edge between groups $g_i$ and $g_j$, if and only if, there is a non-empty set $SP$ of properties in the ontology $O$, and the properties in $SP$ are defined with mapping assertions in $g_i$ and $g_j$. Thus, an edge between $g_i$ and $g_j$ represents that their execution will generate instances of the properties in $SP$ which may overlap and the operator of a duplicate removal is required. 
\item $\phi(g_k,g_q)$ labels an edge between groups $g_i$ and $g_j$ with the set of $SP$ properties that $g_i$ and $g_j$ define in common. \end{itemize}
Figure \ref{fig:exAlgo1}a depicts a graph plan for the grouping named \texttt{Optimized Partition} in Figure \ref{fig:motivating}. The graph is composed of four nodes and one edge, and $\phi(\texttt{Group 2},\texttt{Group 4})$ outputs the set \{\texttt{p3}\} with the property that \texttt{Group 2} and \texttt{Grou} \texttt{p4} both define. 
Initially, Algorithm \ref{alg:bushytree} creates a hyper-node with exactly one group in $GP_{M}$. Figure \ref{fig:exAlgo1}b depicts the initial configuration of the hyper-graph; it is composed of four hyper-nodes. 

Hyper-nodes are sorted in $OL$ based on the degree of connections and the cardinality of the labels of these connections, i.e., the number of properties that the connected groups have in common. Algorithm \ref{alg:bushytree} resorts to this sorting to decide the order in which hyper-nodes will be merged. The first hyper-node $HN$ in $OL$ is selected and combined in a hypernode \textit{NewHN} with the neighbor that shares more properties (\textit{BestNeighbor}). The combined hyper-nodes (i.e., $HN$ and \textit{BestNeighbor}) are eliminated from $OL$ and the new hyper-node (i.e., \textit{NewHN}) is appended at the end of $OL$. If \textit{BestNeighbor} and $HN$ share at least one property in common (i.e., they were connected in the plan graph), \textit{NewHN} is annotated with \texttt{DR} to denote that the duplicate removal needs to be executed. 
This decision implements our heuristic following the optimality principle \textbf{P4}. As a result, duplicate removal is first executed on the union of sets of RDF triples generated by mapping assertions that define the greatest number of properties in common, i.e., an eager evaluation of \texttt{DR} is scheduled.  Contrary, if $HN$ does not have a neighbor, a node with the highest number of connections is selected as best neighbor; \textit{NewHN} is annotated with \texttt{NDR} to denote the union without duplicate removal.
The process is repeated until a fixed point in the hyper-graph is reached; the generated hyper-graph corresponds to the bushy tree. Figures \ref{fig:exAlgo1}b, \ref{fig:exAlgo1}c, \ref{fig:exAlgo1}d, and \ref{fig:exAlgo1}e, illustrate the execution of Algorithm \ref{alg:bushytree}. The generated hyper-graph corresponds to the bushy tree illustrated in Figure \ref{fig:bushyTree}a.  
\begin{figure*}[t!]
    \centering
 \includegraphics[width=1.0\textwidth]{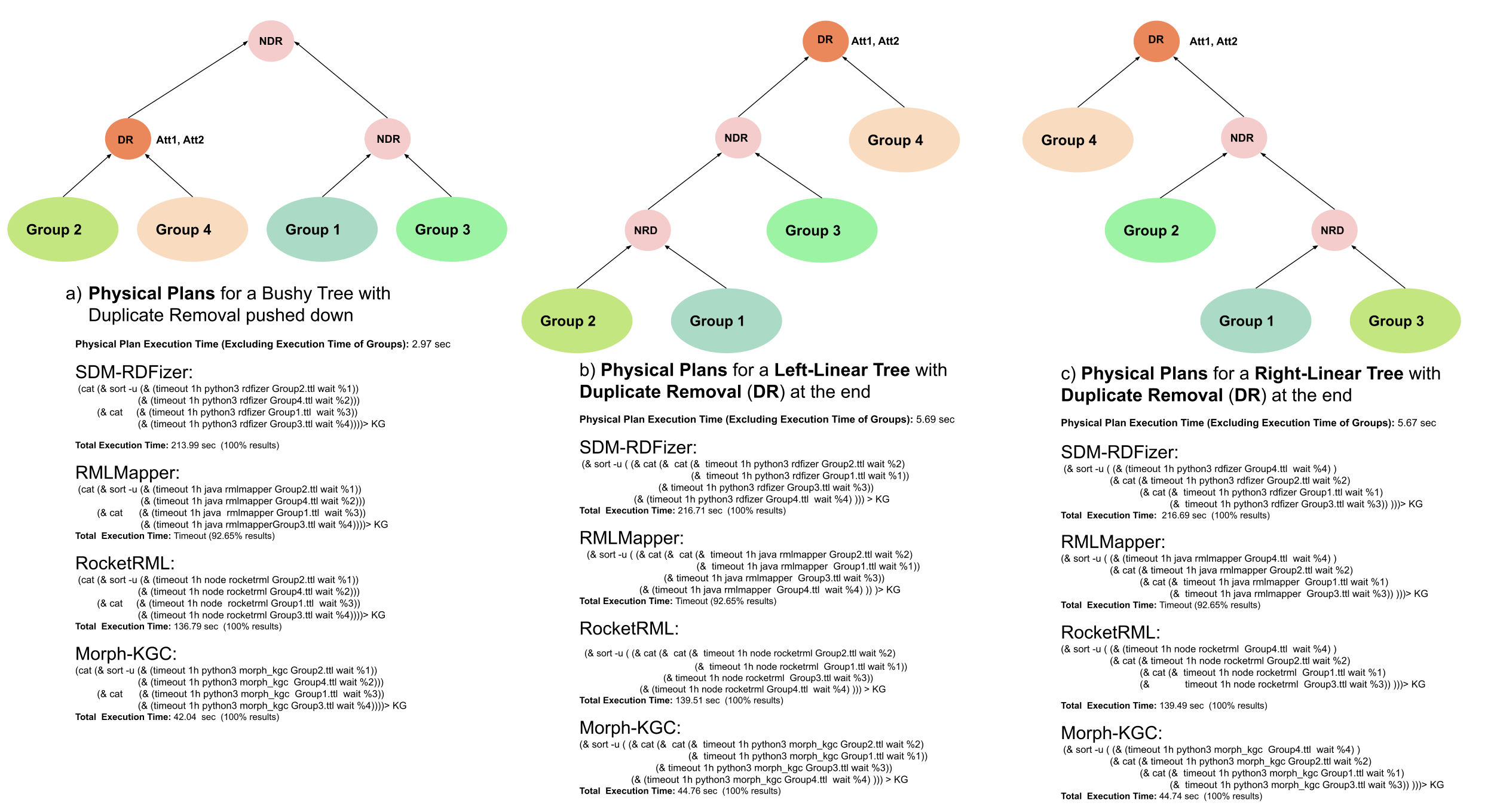}
  \caption{\textbf{Running example.} Physical Plans generated by transforming bushy trees in Figure \ref{fig:bushyTree}. The execution time of the physical plan of the bushy tree (without considering the execution of the groups of assertions) consumes  52.02 \% of the time required for executing the left- and right-linear plans.  }
\label{fig:exAlgo}
\end{figure*}
\begin{theorem}
\label{theo:complexity}
Let $G^{GP_{M}}$ be a graph plan of the groups in $GP_{M}$. Let $n$ be the $G^{GP_{M}}$ cardinality, i.e., the number of groups in $GP_{M}$. The time complexity of Algorithm \ref{alg:bushytree} is $O(n \log n)$ and up to $2^n -1$ bushy sub-plans are generated. 
\end{theorem}

\subsection{Executing Mapping Assertions} 
This step receives a bushy tree $\overline{GP}_{M}$, and generates a physical plan that can execute the mapping assertions in $M$ following the order stated in $\overline{GP}_{M}$. Figure \ref{fig:arquitecture} depicts the main two components of this step of the pipeline. First, nodes in $\overline{GP}_{M}$ are visited following a breadth-first traversal to generate a physical plan. 
A physical plan is defined in terms of operating system commands that enable the execution of a [R2]RML-compliant engine calls to evaluate a group of mapping assertions and generate RDF triples that will be part of a KG.

A physical plan $PP$ is defined as:

 \noindent \textbf{Base Case.} 
Let \textit{ECall}, \textit{Time}, \textit{File}, and \textit{Id} be an [R2]RML engine instruction call, execution timeout, group of mapping assertions file, and $Id$ a process identifier, respectively.
$PP$=\&(timeout \textit{Time} \textit{ECall} wait \%\textit{Id})
represents that \textit{ECall} is executed in the background until the process finalizes or times out after \textit{Time}.

\noindent \textbf{Inductive Case.} 
\begin{itemize}[noitemsep,nolistsep]
    \item \textit{Union with duplicate removal.} Given two physical plans $PP_i$ and $PP_j$ whose execution produces RDF KGs $KG_i$ and $KG_j$ which may overlap. 
    $PP_{i,j}$= \&(sort -u  $PP_i$ $PP_j$) represents that $KG_i$ and $KG_j$ are sorted, merged, and duplicates are removed.
    \item \textit{Union without duplicate removal.}
 Given two physical plans $PP_i$ and $PP_j$ whose execution produces RDF KGs $KG_i$ and $KG_j$ which do not overlap. 
    $PP_{i,j}$= \&(cat  $PP_i$ $PP_j$) represents that $KG_i$ and $KG_j$ are merged.   
    \item \textit{Storing an RDF KG.} Let $PP_i$ be a plan whose execution generates a KG $KG_i$.
    $PP$=$PP_i$ $>$ $KG$ represents that $KG_i$ is stored in the file $KG$.
\end{itemize}

The function $\gamma(\overline{GP}_{M})$ represents the translation of the bushy tree $\overline{GP}_{M}$ into a physical plan $PP$; $\gamma(.)$  is inductively defined over the structure of $\overline{GP}_{M}$ as follows: 

  \noindent \textbf{Base Case.} Let $BT$ be a leaf, i.e., $BT$ is a group of mapping assertions. Let \textit{ECall}, \textit{Time}, \textit{File}, and \textit{BTId} be an [R2]RML engine instruction call, execution timeout, group of mapping assertions file, and $BT$ identifier, respectively. 
  $\gamma(BT)$=(timeout \textit{Time} \textit{ECall} wait \%\textit{BTId})
  \\\noindent \textbf{Inductive Case I}.
Let $BT$ be a binary tree with the operator \texttt{DR} as root node:
\begin{center}
\begin{forest}
[DR
        [BT1]      
        [BT2]
]  
\end{forest}

$\gamma(BT)$=(sort -u \&($\gamma$(BT1)) \&($\gamma$(BT2)))
\end{center}

\noindent \textbf{Inductive Case II}.
Let $BT$ be a binary tree with the operator \texttt{NDR} as root node:
\begin{center}
\begin{forest}
[NDR
        [BT1]      
        [BT2]
]  
\end{forest}

$\gamma(BT)$=(cat \&($\gamma$(BT1)) \&($\gamma$(BT2)))
\end{center}
Figure \ref{fig:exAlgo} depicts the physical plans generated following the inductive definition of $\gamma(.)$. Three different plans are generated: the bushy, left-linear, and right-linear tree plans; the physical plans for each engine are also generated. In these trees, the duplicate removal operator is either pushed down into the tree (Figure \ref{fig:exAlgo} (a)) or in performed at the last step of the evaluation (Figures \ref{fig:exAlgo} (b) and (c)). The execution time of each physical plan is dominated by the cost of evaluating each group of mapping assertions. Nevertheless, the physical plan that implements the bushy tree requires only half of the time (i.e., 52.02\% of the time consumed by the other physical plans) to combine the RDF triples generated during the execution of  \texttt{Group1}, \texttt{Group2}, \texttt{Group3}, and \texttt{Group4}. These results provide evidence of the benefits of scheduling the execution of the KG creation following the physical plans generated by the proposed algorithms.
\begin{table*}[h!]
\resizebox{\linewidth}{!}{
\begin{tabular}{|c|c|l|}
\hline

    \multicolumn{3}{|c|}{\cellcolor[HTML]{7C98AA} Parameter: Dataset Size}\\
    \hline
    \cellcolor[HTML]{AED6F1} Benchmark  & \cellcolor[HTML]{AED6F1} Size &   \cellcolor[HTML]{AED6F1} Description\\ 
\hline
     \multirow{6}{*}{GTFS-Madrid-Bench} & \multirow{2}{*}{1-CSV} & \multirow{2}{25em}{Ten different data sources are 4.8 Mb in total, where SHAPES.csv is the largest file with 4.5 Mb.}\\ 
     && \\ \cline{3-3}
     & \multirow{4}{*}{5-CSV} & \multirow{4}{25em}{Ten different data sources are 10 Mb in total, where SHAPES.csv is the largest file with 7.9 Mb. The KG generated from these data sources is five times bigger than the KG generated from 1-CSV.}\\
     && \\
     && \\
     && \\
\hline 
    \multirow{3}{*}{SDM-Genomic-Datasets} & 10k & Each data source has 10,000 rows.\\ \cline{3-3}
    & 100k & Each data source has 100,000 rows.\\ \cline{3-3}
    & 1M & Each data source has 1,000,000 rows.\\ \cline{3-3}
\hline
    \multicolumn{3}{|c|}{\cellcolor[HTML]{7C98AA} Parameters: Mapping Assertion (MA) Type and Complexity, Selectivity of the Results, and Type of Joins}\\
    \hline
    \cellcolor[HTML]{AED6F1} Benchmark  & \cellcolor[HTML]{AED6F1} Mapping Configuration  &  \cellcolor[HTML]{AED6F1} Description\\ 
\hline 
    \multirow{2}{*}{GTFS-Madrid-Bench}  & \multirow{2}{*}{Standard Config} & \multirow{2}{25em}{13 Concept MAs, 55 Attribute MAs, 73 single-source role MAs, and 12 multi-source role MAs.}\\
    &&\\ 
\hline
    \multirow{19}{*}{SDM-Genomic-Datasets} & Conf1 & One Concept MA, and one Attribute MA.\\ \cline{3-3}
     & Conf2 & One Concept MA, and four Attribute MAs.\\ \cline{3-3}
     & \multirow{2}{*}{Conf3} & \multirow{2}{25em}{Two Concept MA, one referenced-source role MA, and one attribute MA.}\\
     &&\\\cline{3-3}
     & Conf4 & Five Concept MAs, and four Referenced-source role MAs.\\\cline{3-3}
     & Conf5 & Two Concepts MAs, and one Multi-source role MA.\\ \cline{3-3}
     & Conf6 & Five Concept MAs, and Four Multi-source role MAs\\ \cline{3-3}
     & \multirow{2}{*}{AllTogether} & \multirow{2}{25em}{Combines Conf1, Conf2, Conf3, Conf4, Conf5, and Conf6 into one mapping configuration.}\\
     &&\\ \cline{3-3}
     & \multirow{3}{*}{\textbf{Conf7}} & \multirow{3}{25em}{Four Concept MAs, and two Multi-source role MAs. This configuration seeks to evaluate the impact of defining the same predicates using different MAs.}\\ 
     &&\\ 
     &&\\ \cline{3-3}
     & \multirow{4}{*}{\textbf{Conf8}} & \multirow{4}{25em}{Six Concept and five Multi-source role MAs. This mapping configuration aims to recreate a five-star join where five MAs refer to the same parent MA.}\\ 
      &&\\ 
      &&\\ 
     &&\\ \cline{3-3}
     & \multirow{3}{*}{\textbf{Conf9}} & \multirow{3}{25em}{Eight Concept and seven Multi-source role MAs. This configuration combines Conf7 and Conf8 into one mapping configuration.}\\
     &&\\
     &&\\ \cline{3-3}
\hline
\end{tabular}}
\caption{\textbf{Datasets and Configurations of Mapping Assertions.} The table describes each data source and  configuration of MAs used in the experiments and their corresponding benchmarks. Configuration of MAs in bold are considered complex cases. They include several types of MAs of various complexity. Also, they have complex joins (e.g., five-start joins).}
\label{table:2}
\end{table*}

%%%%%%%%%%%%%%%%%%%%%%%%%%%%%%%%%%%%%%%%
%%        EMPIRICAL EVALUATION        %%
%%%%%%%%%%%%%%%%%%%%%%%%%%%%%%%%%%%%%%%%
\section{Experimental Study}
\label{sec:experiments}
The performance of the solution proposed to the problem of \emph{planning KG creation} is studied in four RML-compliant engines: RMLMapper, RocketRML, SDM-RDFizer, and Morph-KGC. The code is publicly available on GitHub\cite{sdm-planner}.
The empirical evaluation aims at answering the following research questions: 
\begin{inparaenum}[\bf RQ1\upshape)]
    \item How does planning the execution of mapping assertions affect the performance of the state-of-the-art RML-compliant engines during KG creation? 
     \item What is the impact of the type of mapping assertions and volume of the data sources on execution time and memory consumed by engines?
    \item What is the impact in-- execution time and memory consumption--of the execution of the mapping assertions following  physical plans generated from bushy trees generated by Algorithm \ref{alg:bushytree}?
\end{inparaenum}

\subsection{Experimental Configuration}
The following setting is configured to assess our research questions. 

\paragraph{Benchmarks}
Experiments are executed on datasets from \textbf{GTFS-Madrid-Bench} and \textbf{SDM-Genomic-Datasets}. Thus, our experimental setting can cover a larger spectrum of parameters that affect a KG creation task, i.e., dataset size, mapping assertion type and complexity, selectivity of the results, and types of joins between mapping assertions.  \autoref{table:2} summarizes the main characteristics of these benchmarks and the covered parameters. 
\\\noindent The \textbf{GTFS-Madrid-Bench}~\cite{chaves2020gtfs} benchmark enables the generation of different configurations of data integration systems whose characteristics impact on the process of KG creation. We generate four logical sources with the scaling factor 1-csv, 5-csv, 10-csv, and 50-csv. The scale value indicates that the comparison between the sizes of the goal KGs. For instance, a KG generated from 5-csv is five times larger than the KG that is created from 1-csv. The logical sources for the 1-csv configuration has in total 4.8 MB. In overall, we consider mapping rules comprised of 13 concept mapping assertions, 55 attribute mapping assertions, 73 single-source role mapping assertions, and 12 multi-source role mapping assertions involving ten data sources.
\\\noindent
\noindent\textbf{SDM-Genomic-Datasets}~\cite{sdm-genomic} is a benchmark to compare the performance of state-of-the-art RML-compliant engines. SDM-Genomic-Datasets is created by randomly selecting data records from somatic mutation data collected in COSMIC~\cite{cosmic}. SDM-Genomic-Datasets includes eight different logical data sources with various sizes including 10k, 100k, 1M, and 10M number of rows. Accordingly, every pair of logical data sources with the same size differ in data duplicate rates, which can be either 25\% or 75\%. Each duplicate value is repeated 20 times. For example, a 10k logical data source with 25\% data duplicate rates has 75\% duplicate-free records (i.e., 7,500 rows) and the rest of the 25\% of the records (i.e., 2,500 rows) correspond to 125 different records which are duplicated 20 times. The SDM-Genomic-Datasets offers nine  mapping assertion configurations. \begin{inparaenum}[\bf Conf1\upshape :]
    \item Set of two mapping assertions with one concept and one attribute mapping assertions.
    \item Set of five mapping assertions, including one concept and four attribute mapping assertions. 
    \item Set of four mapping assertions consisting of two concepts, one referenced-source role, and one attribute mapping assertions.
    \item Set of nine mapping assertions with five concepts and four referenced-source role mapping assertions.
    \item Set of three mapping assertions comprised of two concepts and one multi-source role mapping assertions.
    \item Set of nine mapping assertions, including five concepts and four multi-source role mapping assertions.
\end{inparaenum}
We group the aforementioned mapping assertions into a set named \textbf{AllTogether}. Furthermore, the benchmark includes three extra configurations to enable the evaluation of the impact of two other influential parameters on the performance of KG creation frameworks \cite{sdm-genomic-mapping}.
%The first configuration, 
\textbf{Conf7} aims at evaluating the impact of defining the same predicates using different mapping assertions. \textbf{Conf8} provides a mapping rule which is connected to five other mapping rules with different logical sources through join, i.e., this mapping assertion is connected via a five-star join with the other five mapping assertions. The last configuration or \textbf{Conf9} combines the first two configurations in one testbed. \textbf{Conf7}: Set of four mapping assertions with four concepts and two multi-source role mapping assertions. For each pair of mapping assertions, there is a multi-source role mapping assertion. The data sources of one pair of the mapping assertions are a subset of the other pair. Both pairs of mapping assertions share the same predicate. \textbf{Conf8}: Set of six mapping assertions with six concepts and five multi-source role mapping assertions. In this set, five child mapping assertions are referring to the same parent mapping assertion. \textbf{Conf9}: Set of eight mapping assertions with eight concepts and seven multi-source role mapping assertions. 

\begin{figure*}[t!]
\centering
    \begin{subfigure}{.7\linewidth}
    \centering
        \includegraphics[width=1.0\linewidth]{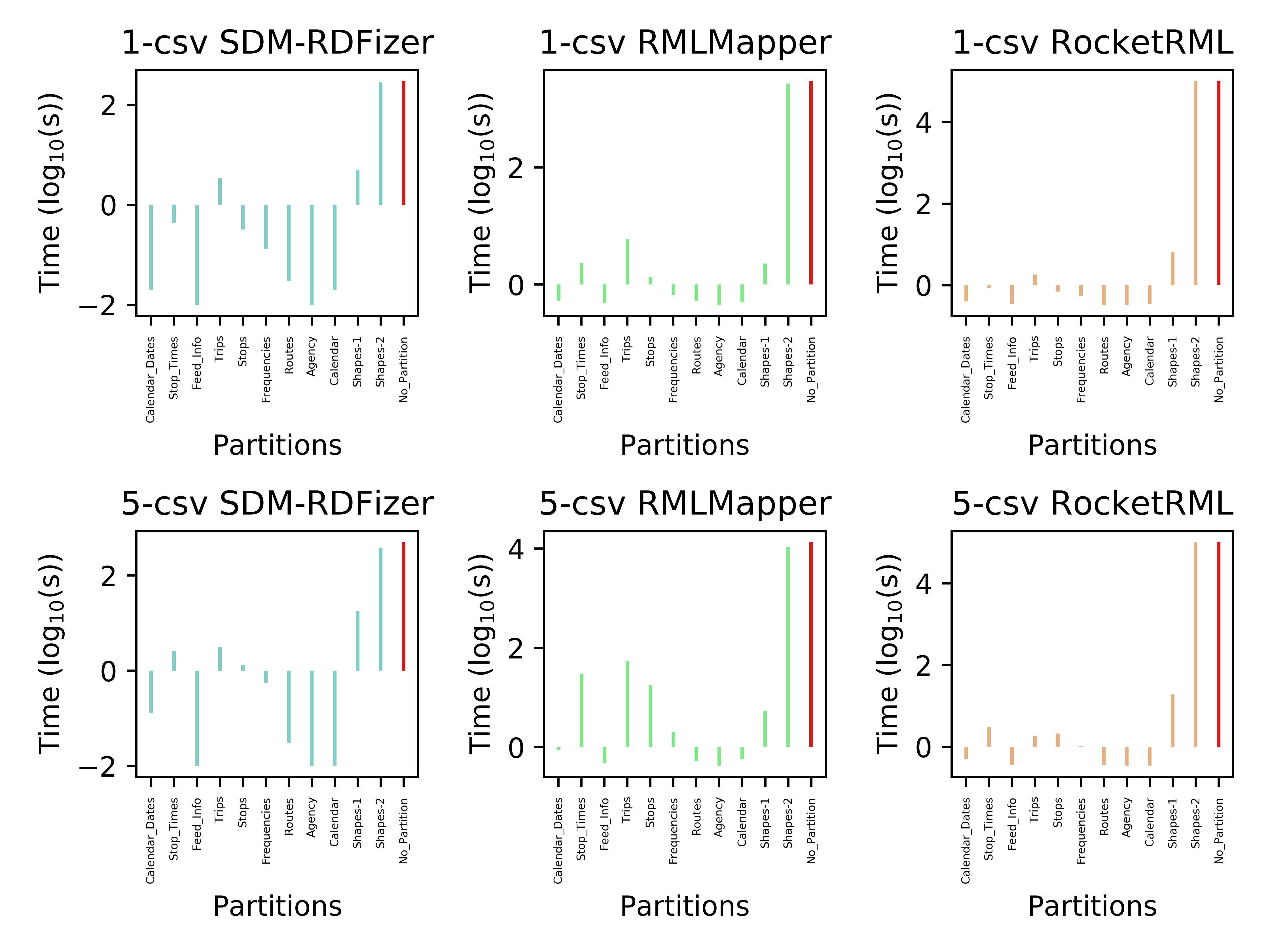}
   \vspace{-0.25cm}
    \caption{Planning Impact on Execution Time }
    \label{fig:time_execution}
    \end{subfigure}
     \begin{subfigure}{0.7\linewidth}
    \centering
        \includegraphics[width=1.0\linewidth]{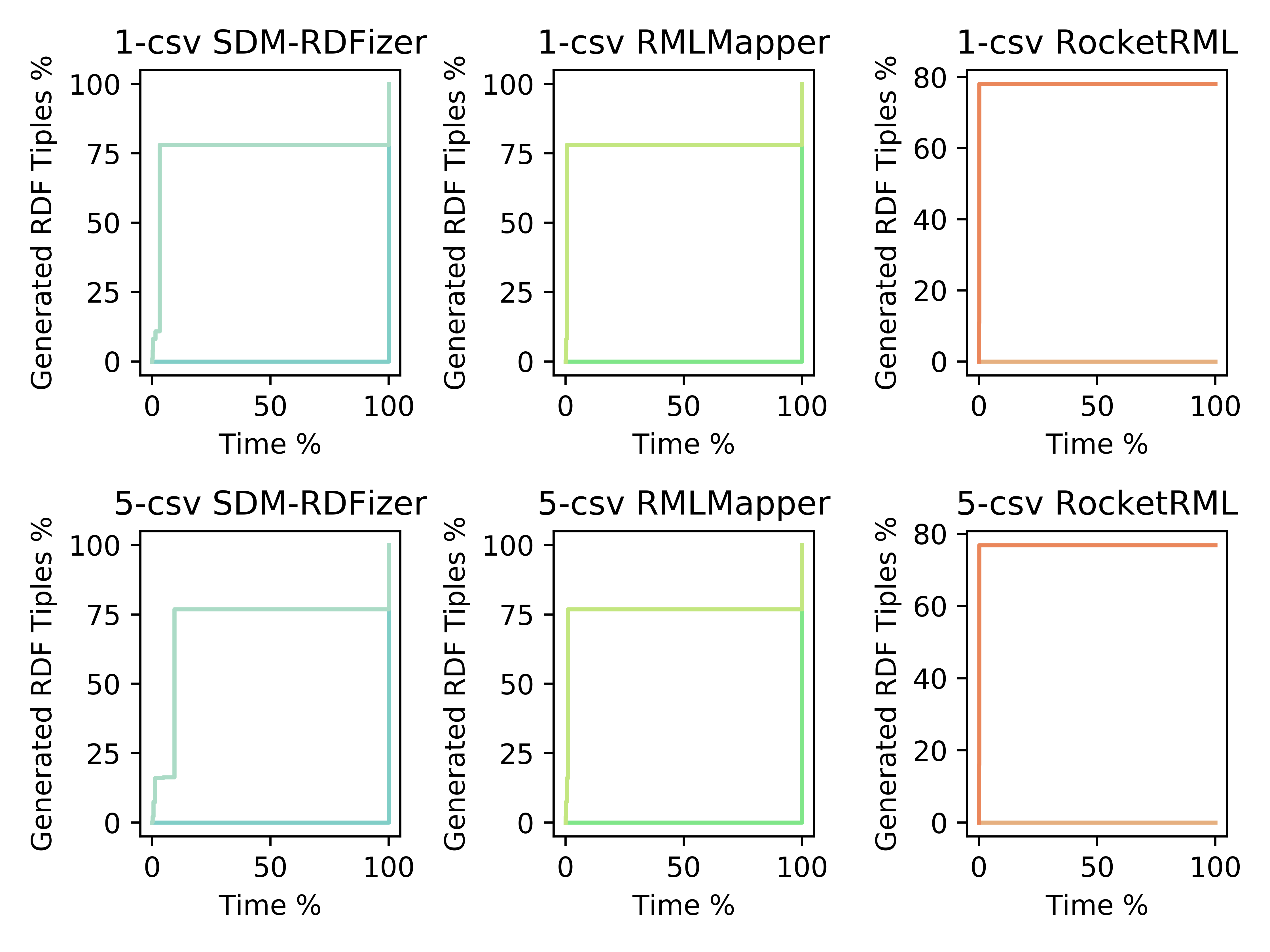}
   \vspace{-0.25cm}
    \caption{Planning Impact on Execution Time and KG Completeness}
    \label{fig:time_execution_complete}
    \end{subfigure}
    \caption{\textbf{Planning Impact on the GTFS-Madrid-Bench.} The effects of the proposed planning techniques over the GTFS-Madrid-Bench data sources: 1-csv, and 5-csv. SDM-RDFizer v3.6, RMLMapper, and RocketRML. Figure a presents the execution time of each individual partition and the entire mapping. We can observe that the Shapes-2 partition takes the longest time among the partitions. Figure b illustrates the percentage of RDF triples that are generated over the percentage of time. For RocketRML, since it was capable of executing the Shapes-2 partition was only able to generate approximately $80\%$ of the KG}
    \label{fig:time}
\end{figure*}
\iffalse
\begin{figure*}[t]
    \centering
    \begin{subfigure}{.35\linewidth}
    \centering
        \includegraphics[width=1.0\linewidth]{images/1-csv_time.png}
   \vspace{-0.25cm}
    \caption{Execution Time in 1-csv}
    \label{fig:time:1-csv}
    \end{subfigure} \hspace{-0.6cm}
     \begin{subfigure}{.35\linewidth}
      \centering
        \includegraphics[width=1.0\linewidth]{images/5-csv_time.png}
        \vspace{-0.25cm}
         \caption{Execution Time in 1-csv}
        \label{fig:time:5-csv}
\end{subfigure} 
\begin{subfigure}{.25\textwidth}     
\resizebox{1.0\linewidth}{!}{%
\begin{tabular}{|l|c|c|}
\hline
\rowcolor[HTML]{afdbea} 
\textbf{Engine}  & \multicolumn{1}{l|}{\cellcolor[HTML]{afdbea}\textbf{1-csv}} & \multicolumn{1}{l|}{\cellcolor[HTML]{afdbea}\textbf{5-csv}} \\ \hline
\rowcolor[HTML]{eefde9} 
RMLMapper  & 100\% & 100\%  \\ \hline
\rowcolor[HTML]{eefde9} 
RMLMapper+Planning  & TTT   & XXXX \\\hline
\rowcolor[HTML]{eefde9} 
RochetRML   & TTT & PPP  \\ \hline
\rowcolor[HTML]{eefde9} 
RochetRML+Planning  & XXX     &XXX \\\hline
\rowcolor[HTML]{eefde9} 
SDM-RDFizer & ZZZZ  & ZZZ \\ \hline
\rowcolor[HTML]{eefde9} 
SDM-RDFizer+Planning & YYY    & XXX \\ \hline
\end{tabular}
}
 \caption{Percentage RDF triples}
        \label{fig:rdfTriples}
\end{subfigure}
\caption{Execution time of each KG Creation Engine in each Dataset.}
    \label{fig:time}
\end{figure*}
\fi

\paragraph{RML Engines}
RMLMapper v4.12 \cite{rmlMapper},  RocketRML v1.11.3 \cite{rocketrml}, Morph-KGC v1.4.1 \cite{morphKGC}, and
SDM-RDFizer v3.6 \cite{sdm-rdfizer3.6}. Recently, SDM-RDFizer v4.0 \cite{sdm-rdfizer4.0} has been published. According to the tool description, SDM-RDFizer v4.0 implements planning techniques, physical operators for the execution of mapping assertions, and data compression techniques for reducing the size of the main memory structures required to store intermediate results. In order to create a fair evaluation of the performance of the techniques developed in SDM-RDFizer v4.0, we implement an upgraded version of SDM-RDFizer v3.6 which includes the data compression technique developed in SDM-RDFizer v4.0; we call this engine SDM-RDFizer v4.0$^{--}$.     

\paragraph{Implementations.}
The planning and execution pipeline is implemented in Python 3. The compression techniques implemented in SDM-RDFizer v4.0$^{--}$ encode RDF resources generated during the KG creation process. For each RDF resource \textit{R}, an identification number \textit{i} is assigned to it. Thus, RDF triples are built not from RDF resources but the identification number. Moreover, each identification number \textit{i} is encoded in Base36 to reduce the memory usage further. Base36 is an encoding scheme that transforms a string into a 36 characters representation. The characters used are the letters from A to Z and the numbers from 0 to 9. For example, the number "95634785" is encoded as "1KXS9T". The SDM-RDFizer operators are adapted to consider this compression method, consuming less main memory. 

\paragraph{Metrics} We consider two metrics to evaluate the efficiency of our proposed approach. \textit{Execution time} is defined as the elapsed time required to generate the bushy tree and execute the corresponding physical plan used to create the KG. It is measured as the absolute wall-clock system time, as reported by the \verb|time| command of the Linux operating system. The leaves of a bushy tree are executed in parallel, and execution of the leaves corresponds to the greatest execution time; execution time also includes the time of merging the results generated during the execution of the tree leaves. \textit{Memory consumption} is determined as the amount of memory that is consumed during the generation of a KG. The memory usage is measured by using the \texttt{tracemalloc} library from Python \cite{malloc}. The \texttt{get\_traced\_memory()} method from \texttt{tracemalloc} returns the amount of memory currently being used. This method presents the memory usage in Kilobytes, for ease of use, it is converted into Megabytes. The timeout is five hours. The experiments are executed in an Intel(R) Xeon(R) equipped with a CPU E5-2603 v3 @ 1.60GHz 20 cores, 64GB memory and with the O.S. Ubuntu 16.04LTS. 
All the resources used in the reported experimental study are publicly available \cite{planner4KG}.
\begin{figure*}[t!]
    \centering
    \subfloat[10k records with 25\% duplicate rate.]{
        \includegraphics[trim=0 0 0 21,clip,width=0.33\linewidth]{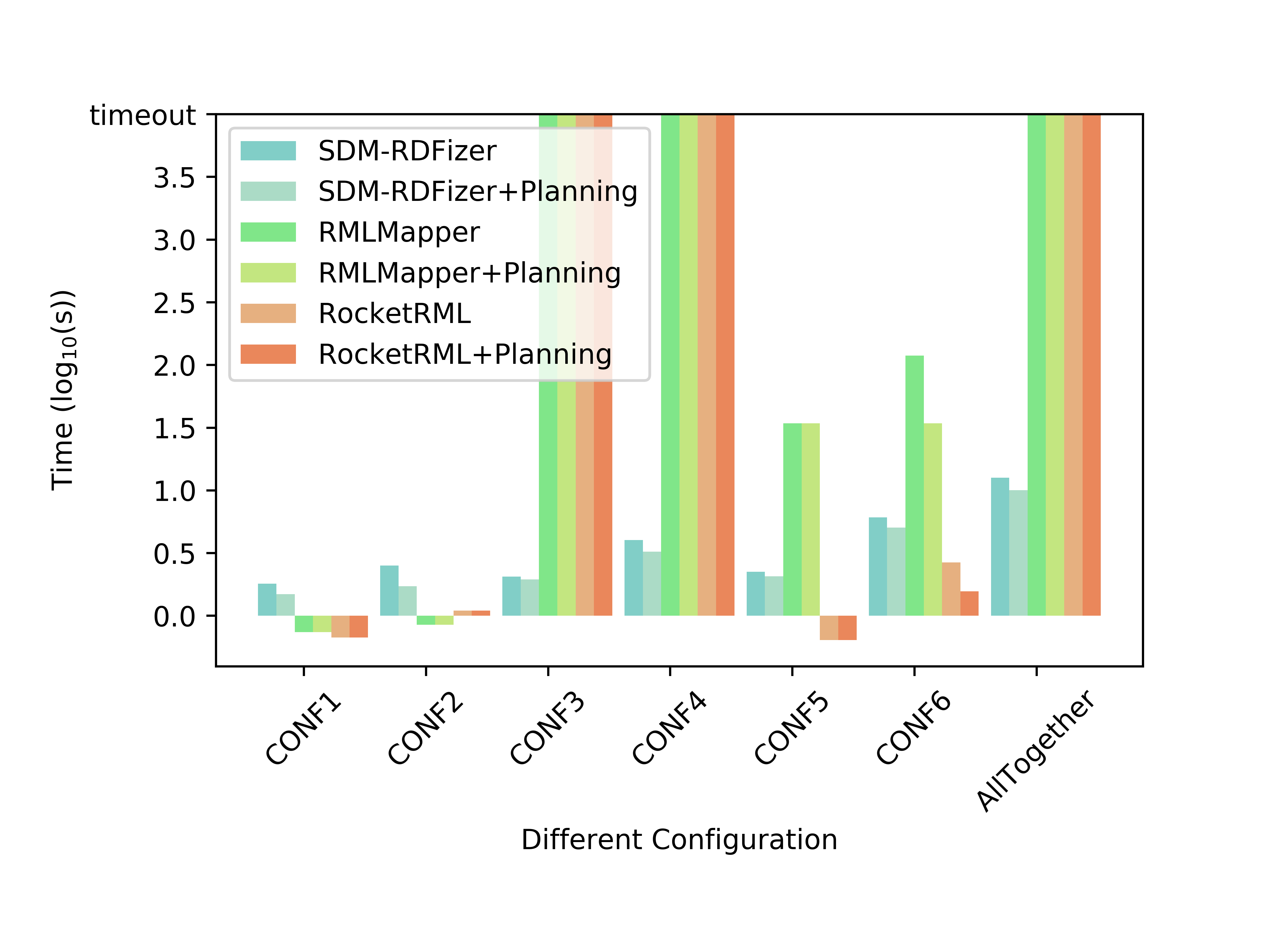}
        \label{fig:genomic:10k_25}
    }
    \subfloat[100k records with 25\% duplicate rate.]{
        \includegraphics[trim=0 0 0 21,clip,width=0.33\linewidth]{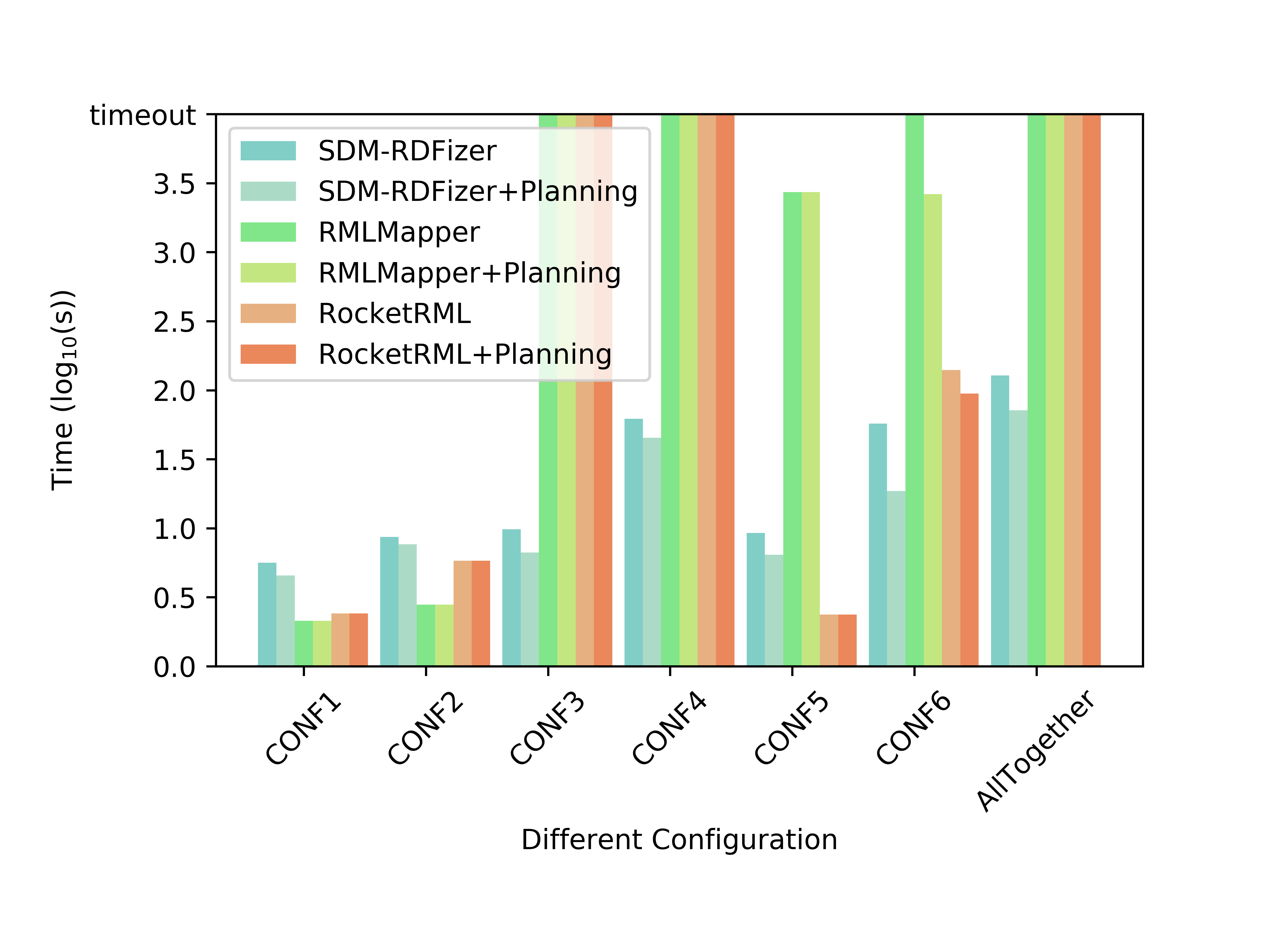}
        \label{fig:genomic:100k_25}
    }
    \subfloat[1M records with 25\% duplicate rate.]{
        \includegraphics[trim=0 0 0 21,clip,width=0.33\linewidth]{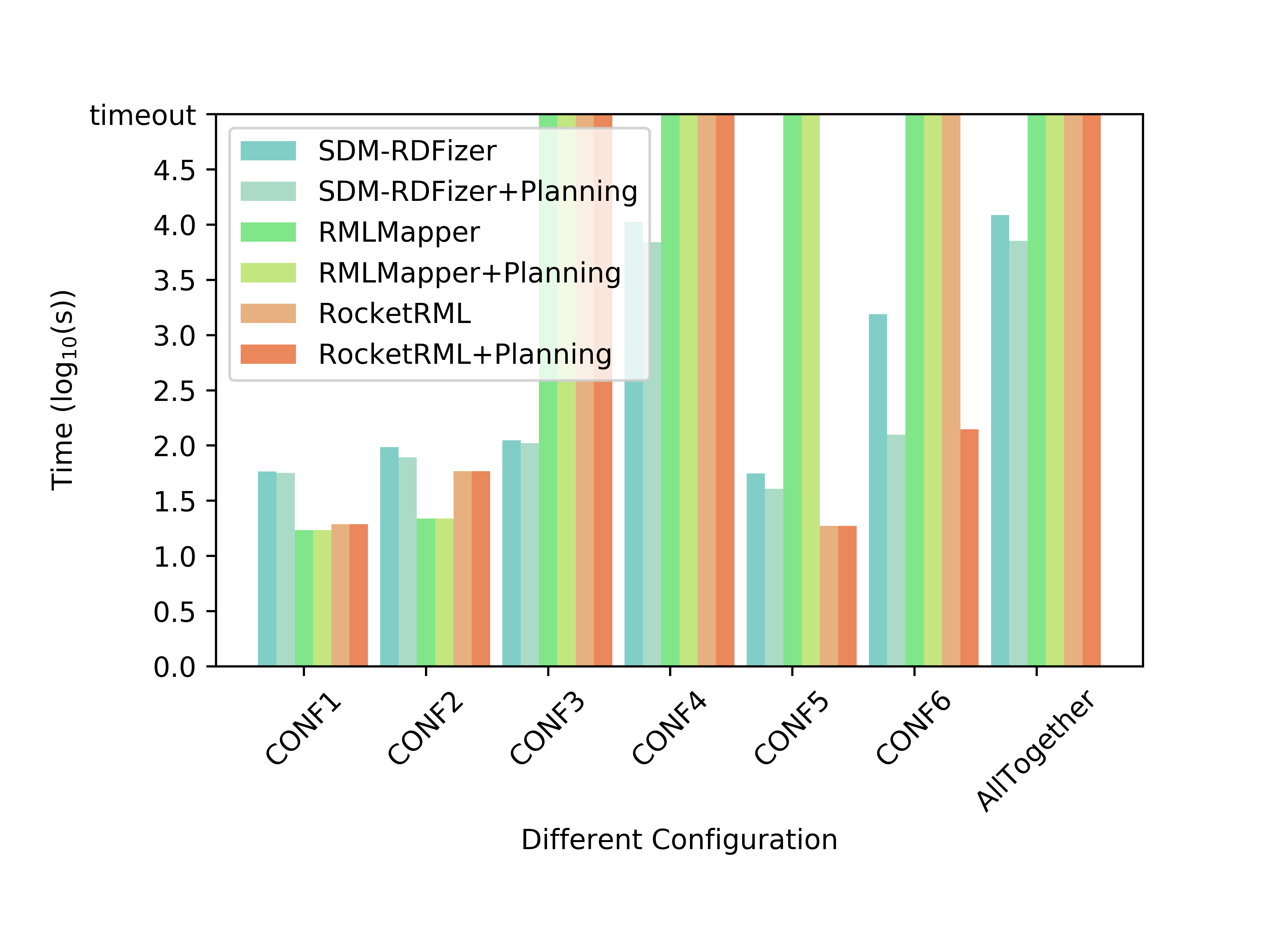}
        \label{fig:genomic:1M_25}
    }
    ~\\\vspace*{.75em}
    \subfloat[10k records with 75\% duplicate rate.]{
        \includegraphics[trim=0 0 0 21,clip,width=0.33\linewidth]{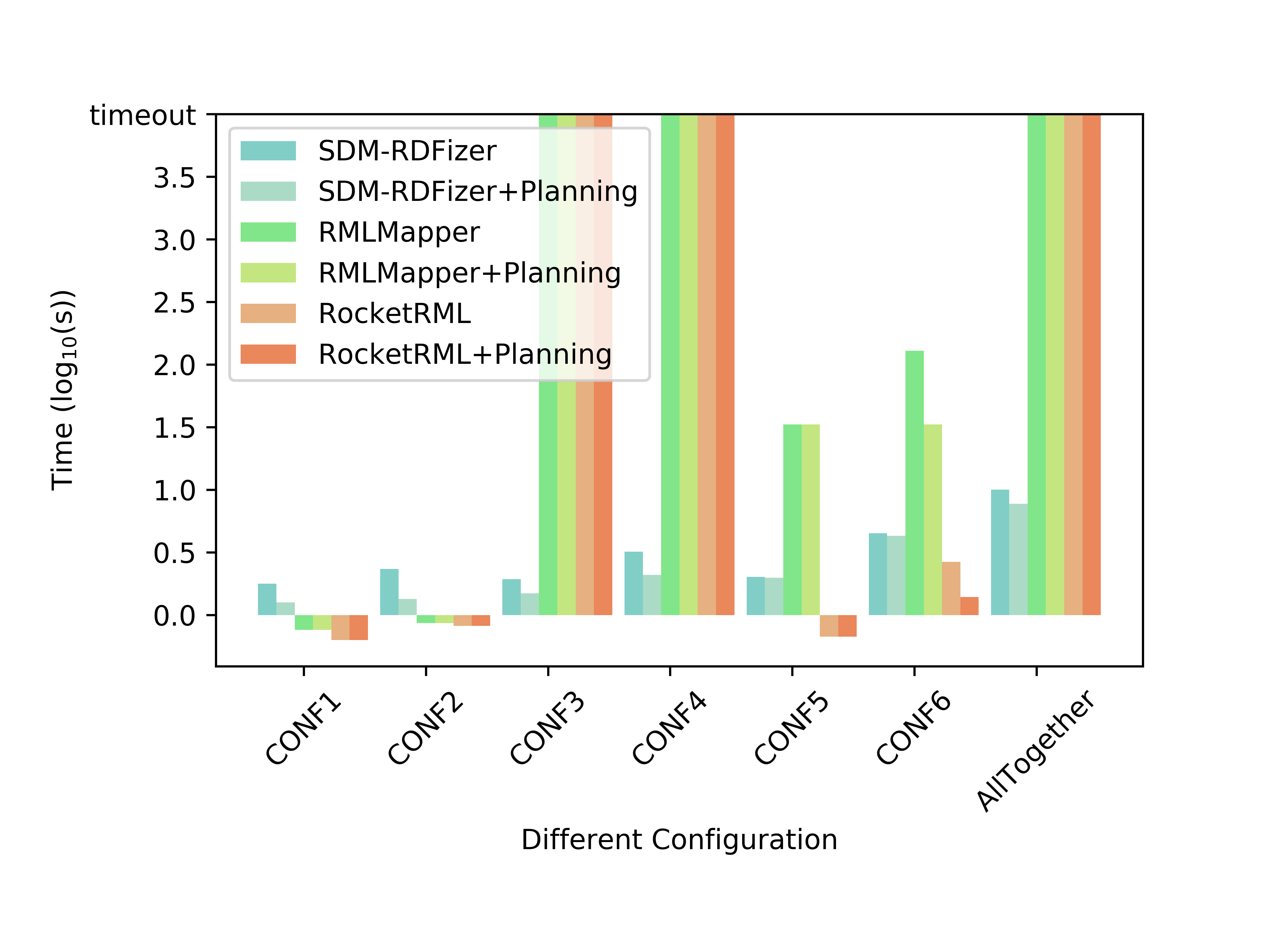}
        \label{fig:genomic:10k_75}
    }
    \subfloat[100k records with 75\% duplicate rate.]{
        \includegraphics[trim=0 0 0 21,clip,width=0.33\linewidth]{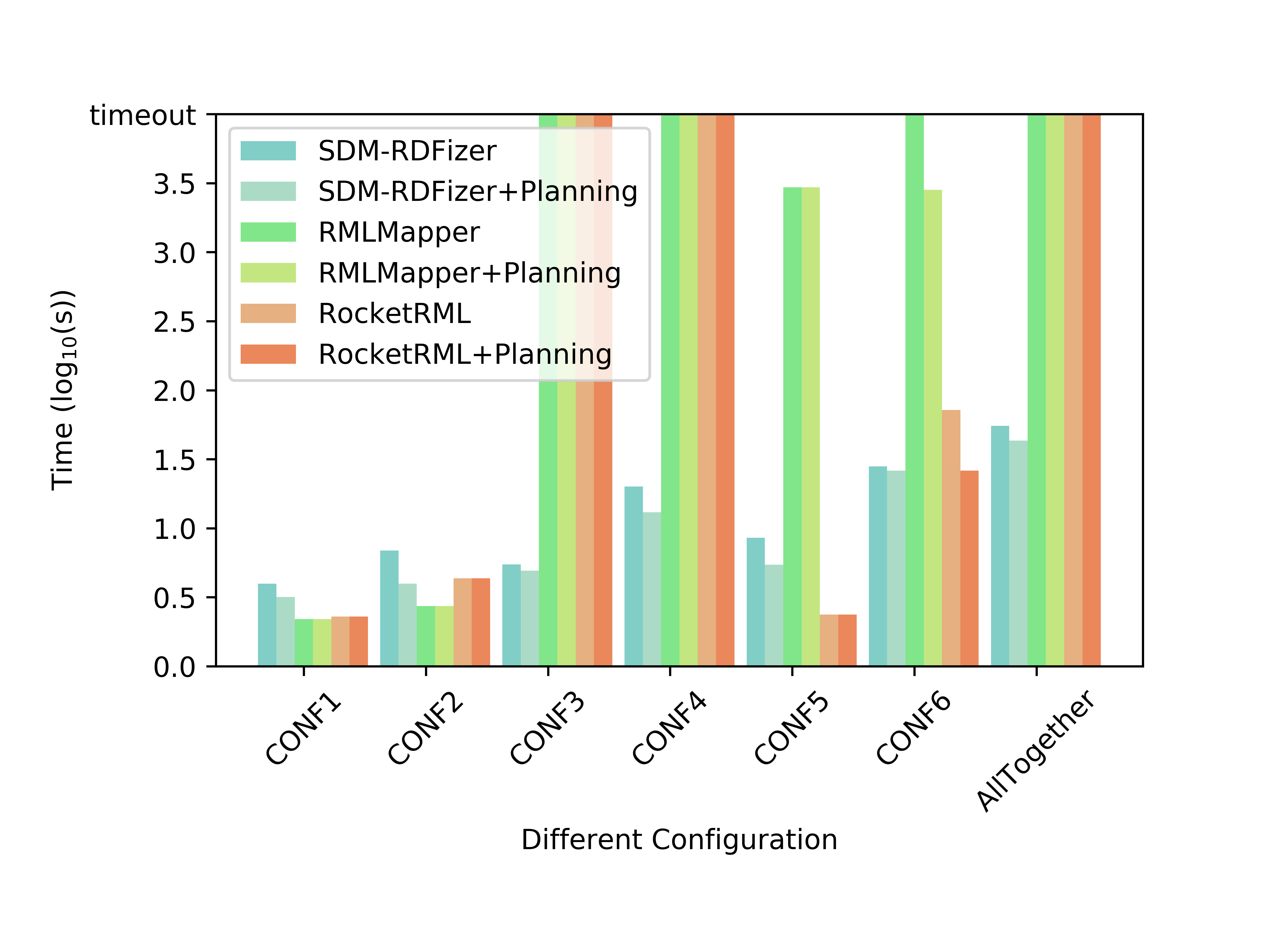}
        \label{fig:genomic:100k_75}
    }
    \subfloat[1M records with 75\% duplicate rate.]{
        \includegraphics[trim=0 0 0 21,clip,width=0.33\linewidth]{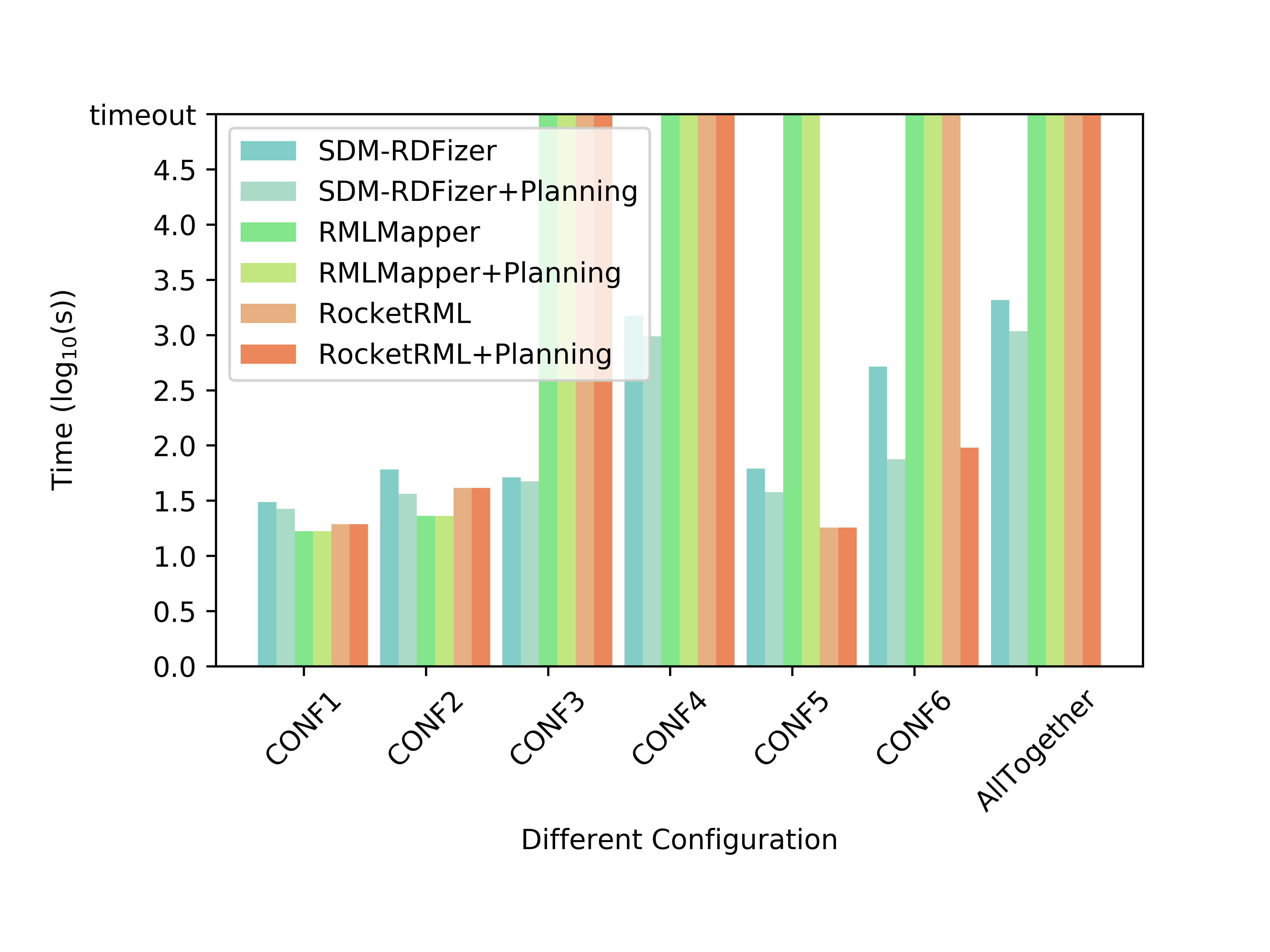}
        \label{fig:genomic:1M_75}
    }
    \caption{\textbf{Results of the Execution of the GENOMIC benchmark.} Execution time of Conf1, Conf2, Conf3, Conf4, Conf5, Conf6, and AllTogether for SDM-RDFizer v3.6, RMLMapper, and RocketRML.}
    \label{fig:genomic}
\end{figure*}

\subsection{Experiment 1- Efficiency on GTFS-Madrid-Bench}
This experiment aims at evaluating the impact that grouping mapping assertions have on the performance of the state-of-the-art engines RMLMapper, RocketRML, and SDM-RDFizer v3.6. Using the algorithm of \emph{Grouping Mapping Assertions}, ten groups of mapping assertions are generated, which are evaluated over the 1-csv and 5-csv data sources from GTFS-Madrid-Bench. Moreover, the full set of mapping assertions is executed by each engine considering both data sources. Figure \ref{fig:time} reports on the execution time (seconds in log scale) for each of the ten groups, as well as with \texttt{No\_Partition}. The three engines can execute nine groups in less than five seconds and produce 15.63$\%$ of the overall RDF triples. 
\\\noindent The group \texttt{Shapes} comprises four attribute mapping assertions and a multi-source role mapping assertion partition over one logical source named ``Shapes", i.e., the execution of this assertion requires a self-join. We further divided the group \texttt{Shapes} into two smaller partitions: \texttt{Shapes-1} containing the four attribute mapping assertions and \texttt{Shapes-2} containing the self-join. 
\\\noindent
We generate these smaller partitions because RMLMapper and RocketRML cannot complete the execution of the group \texttt{Shapes}.
\\\noindent The size of the logical source ``Shapes" is 4.5MB in the case of 1-csv and 7.9MB in 5-csv. RocketRML is unsuccessful in finishing the evaluation of the self-join due to memory failure. Contrary, RMLMapper, and SDM-RDFizer succeed to execute this group of mapping assertions over the two studied versions of the data source ``Shapes" (Figures \ref{fig:time_execution} and \ref{fig:time_execution_complete}). RMLMapper produces the overall RDF triples of the ``Shapes" in 2,707.32 seconds and 10,800.32 seconds in case of 1-csv and 5-csv, respectively. SDM-RDFizer also generates all the RDF triples of the ``Shapes" in 284.06 seconds and 396.2 seconds for 1-csv and 5-csv, respectively. 
\\\noindent
In the case of \texttt{No\_Partition}, RocketRML runs out of memory without generating any result, while RMLMapper and SDM-RDFizer both produce all the RDF triples. 
In the optimized case, i.e., the time of executing the groups of assertions in parallel, RMLMapper requires, respectively, 91.42$\%$ and 80.87$\%$ in 1-csv and 5-csv of the time \texttt{No\_Partition}.
\\\noindent
Likewise, the proposed planning techniques also speed up the SDM-RDFizer execution concerning \texttt{No Partition}; it consumes, respectively, 96.32$\%$ and 79.50$\%$ in 1-csv and 5-cvs of the execution time of \texttt{No\_Partition}.
Although savings are observed, the evaluation of the \texttt{Shapes} group consumes the majority of the execution time of the corresponding physical plan. This prevents observing the benefits of executing the mapping assertions in parallel. 
\\\noindent
It is also important to highlight that even though this benchmark, allows for configuring testbeds that produce KGs of various sizes, the scaling factor is not equally applied to all the data sources and RDF triples produced by each mapping assertion.
Conversely, most of the new RDF triples produced by a high-scaled KG are generated by the \texttt{Shapes} group. This lack of diversity also prevents observing differences in different configurations, i.e., 1-csv and 5-csv.
\subsection{Experiment 2- Efficiency on SDM-Genomic-Datasets}
This experiment aims to assess the impact of planning on a real-world dataset such as the one provided by the SDM-Genomic-Datasets. Although the mapping assertions defined for the SDM-Genomic-Datasets are much simpler compared to the ones in GTFS-Madrid-Bench, they cover all the different types of mapping assertions presented in \autoref{sec:preliminaries}. 
\begin{figure*}[h!]
    \centering
  \begin{subfigure}{.50\linewidth}
  \centering
      \includegraphics[width=0.75\linewidth]{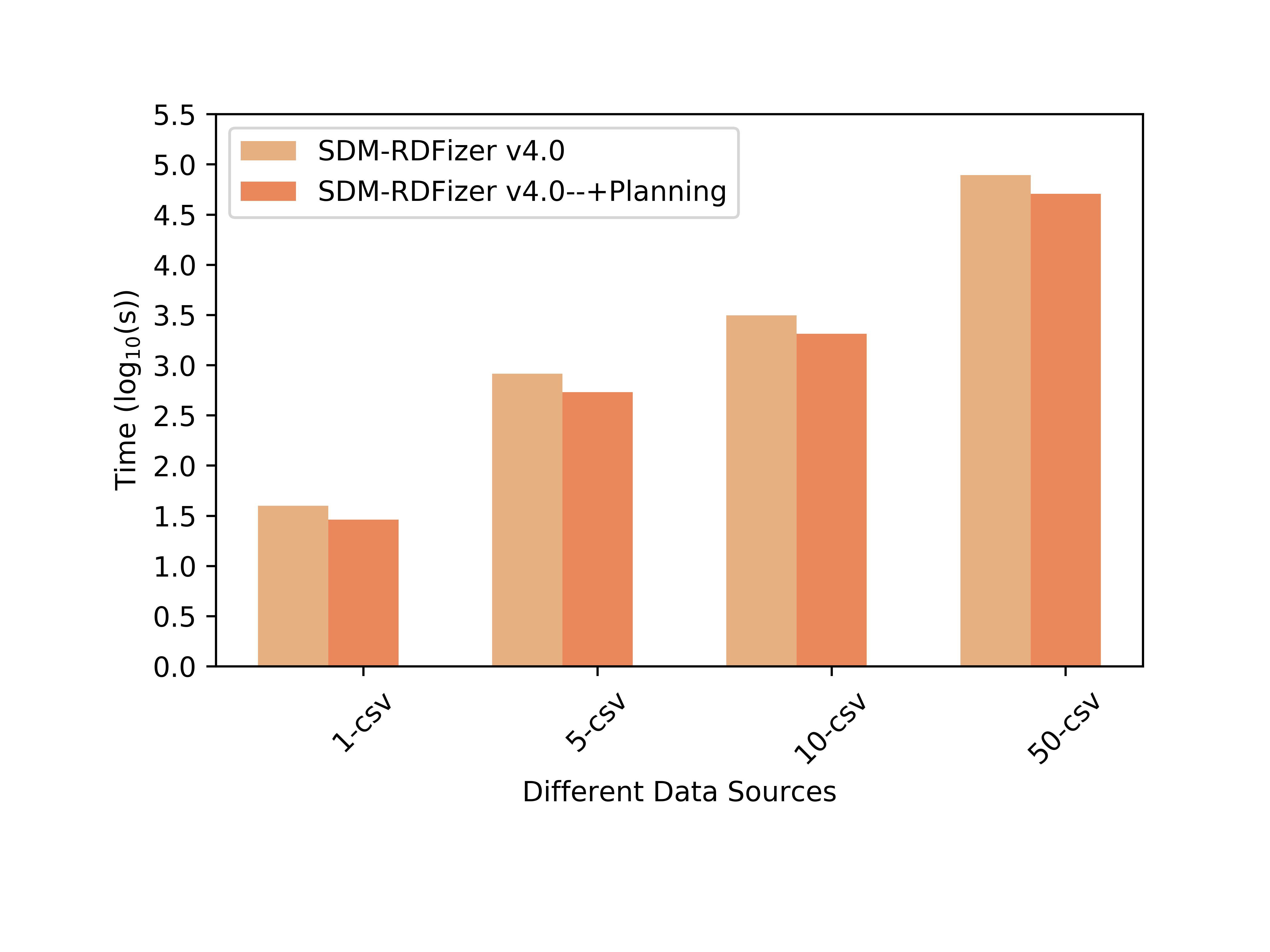}
         \vspace{-0.5cm}
    \caption{Execution Time}
    \label{fig:experiment_time}
    \end{subfigure}\hfill
  \begin{subfigure}{.50\linewidth}
  \centering
      \includegraphics[width=0.75\linewidth]{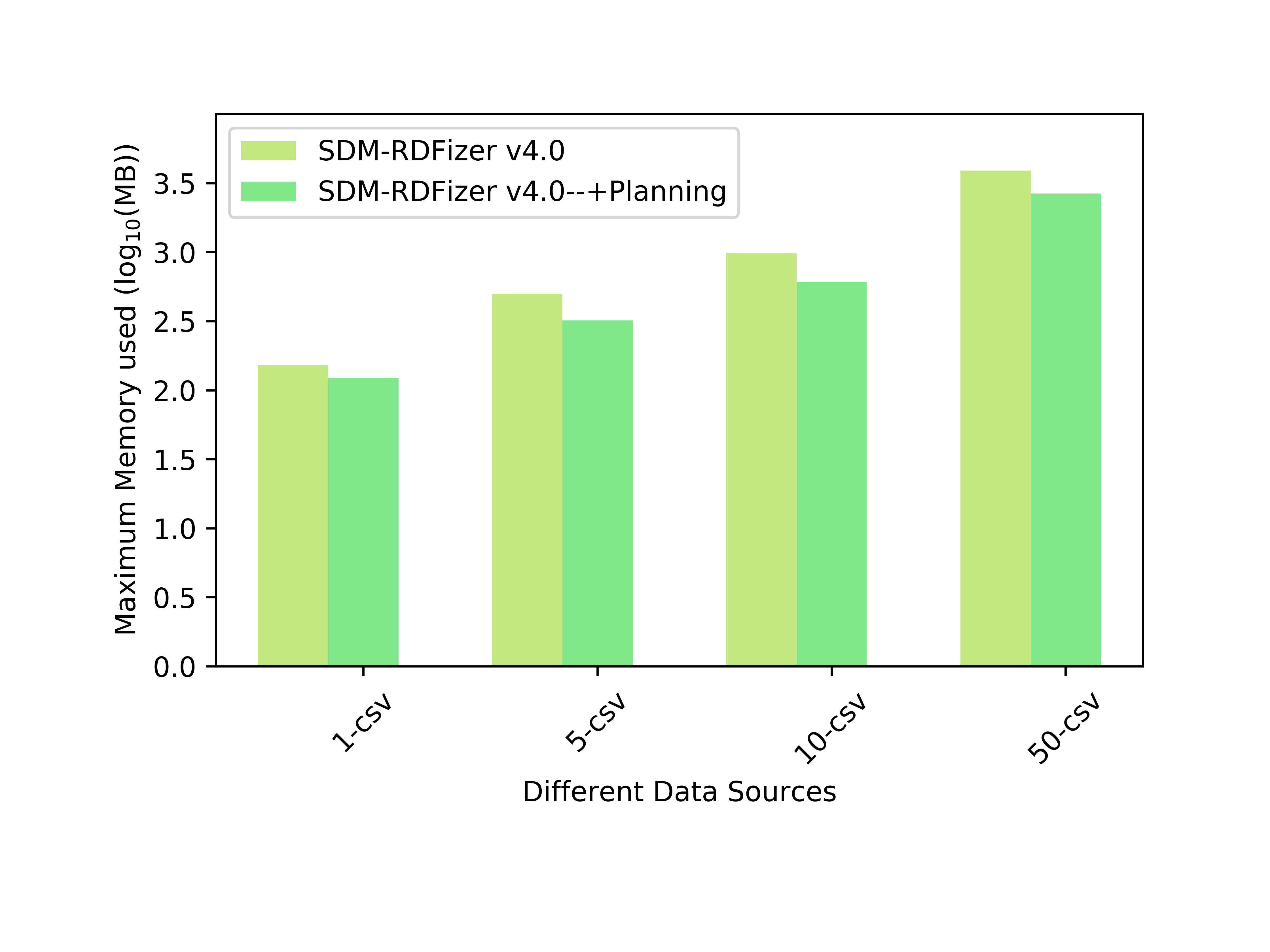}
         \vspace{-0.5cm}
    \caption{Maximum Memory Usage}
    \label{fig:size}
    \end{subfigure}
    \caption{\textbf{Optimize Planning}. The effects of proposed planning techniques over the GTFS-Madrid-Bench data sources: 1-csv, 5-csv, 10-csv, and 50-csv. SDM-RDFizer v4.0$^{--}$+Planning, and SDM-RDFizer 4.0}
    \label{fig:experiment3}
\end{figure*}
We study the performance of each engine, i.e., RocketRML, RMLMapper, and SDM-RDFizer in presence and absence of planning using SDM-Genomic-Datasets. 
In addition to the six configurations of mapping assertions, i.e., \textbf{Conf1}, \textbf{Conf2}, \textbf{Conf3}, \textbf{Conf4}, \textbf{Conf5}, and \textbf{Conf6},  we consider an additional configuration consisting of the union of all them. 
\begin{figure*}[h!]
    \centering
    \includegraphics[width=\linewidth]{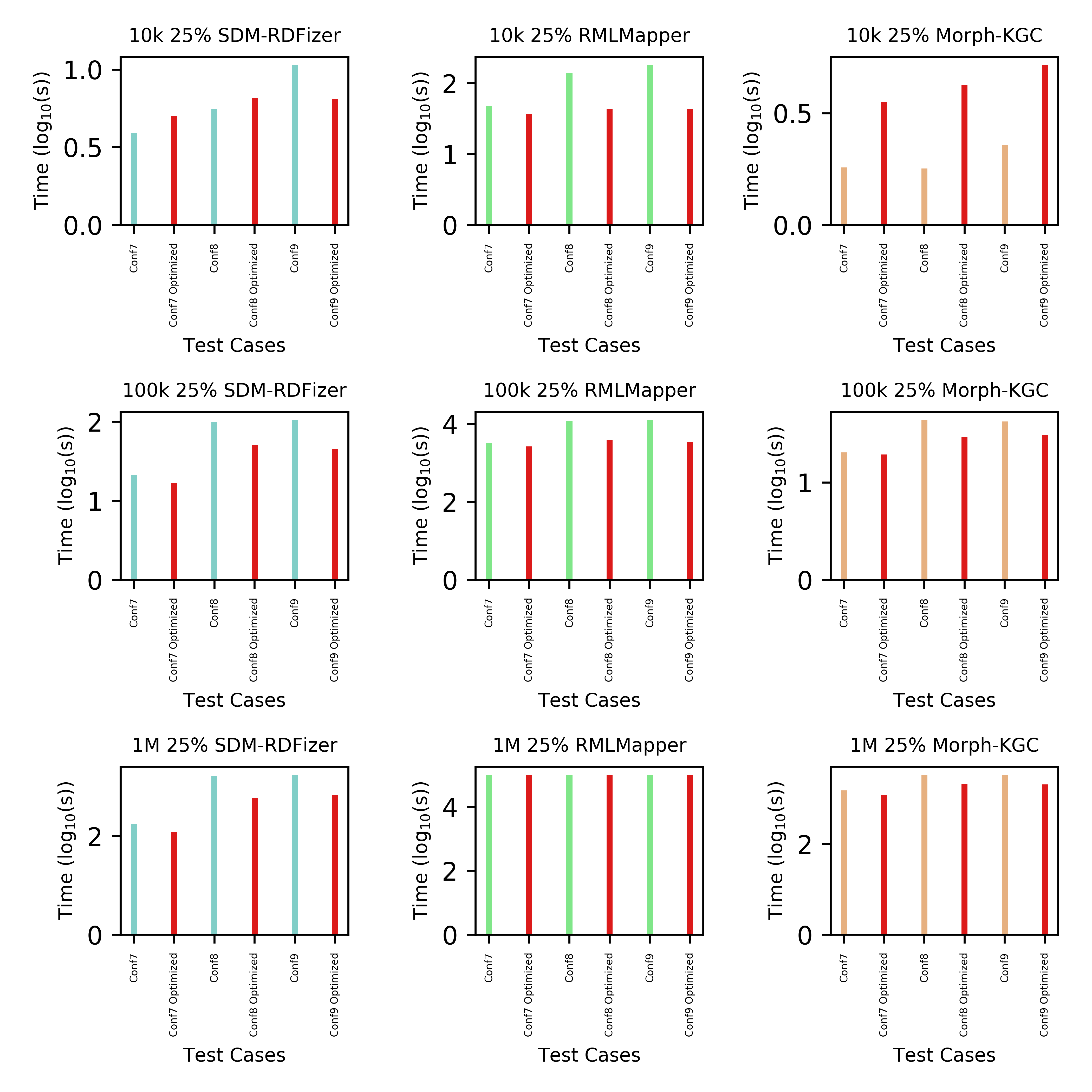}
    \caption{\textbf{Efficiency Planning For Complex Cases with 25\% duplicate rate}. The effects of proposed planning techniques over the SDM-Genomic-Datasets with 25\% duplicate rate over Conf7, Conf8, and Conf9. SDM-RDFizer v3.6+Planning, RMLMapper+Planning, Morph-KGC+Planning}
    \label{fig:complex_25}
\end{figure*}
We refer to it as \texttt{AllTogether}. As illustrated in \autoref{fig:genomic}, in the case of having referenced-source role mapping assertions (i.e., \textbf{Conf3} and \textbf{Conf4}), neither of the two engines, RMLMapper and RocketRML, is able to complete the execution before the timeout. As observed in \autoref{fig:genomic}, applying planning in simple cases like \textbf{Conf1}, \textbf{Conf2}, and \textbf{Conf3} with low data duplicate rates does not show a considerable impact on the performance. 
\begin{figure*}[t!]
    \centering
    \includegraphics[width=\linewidth]{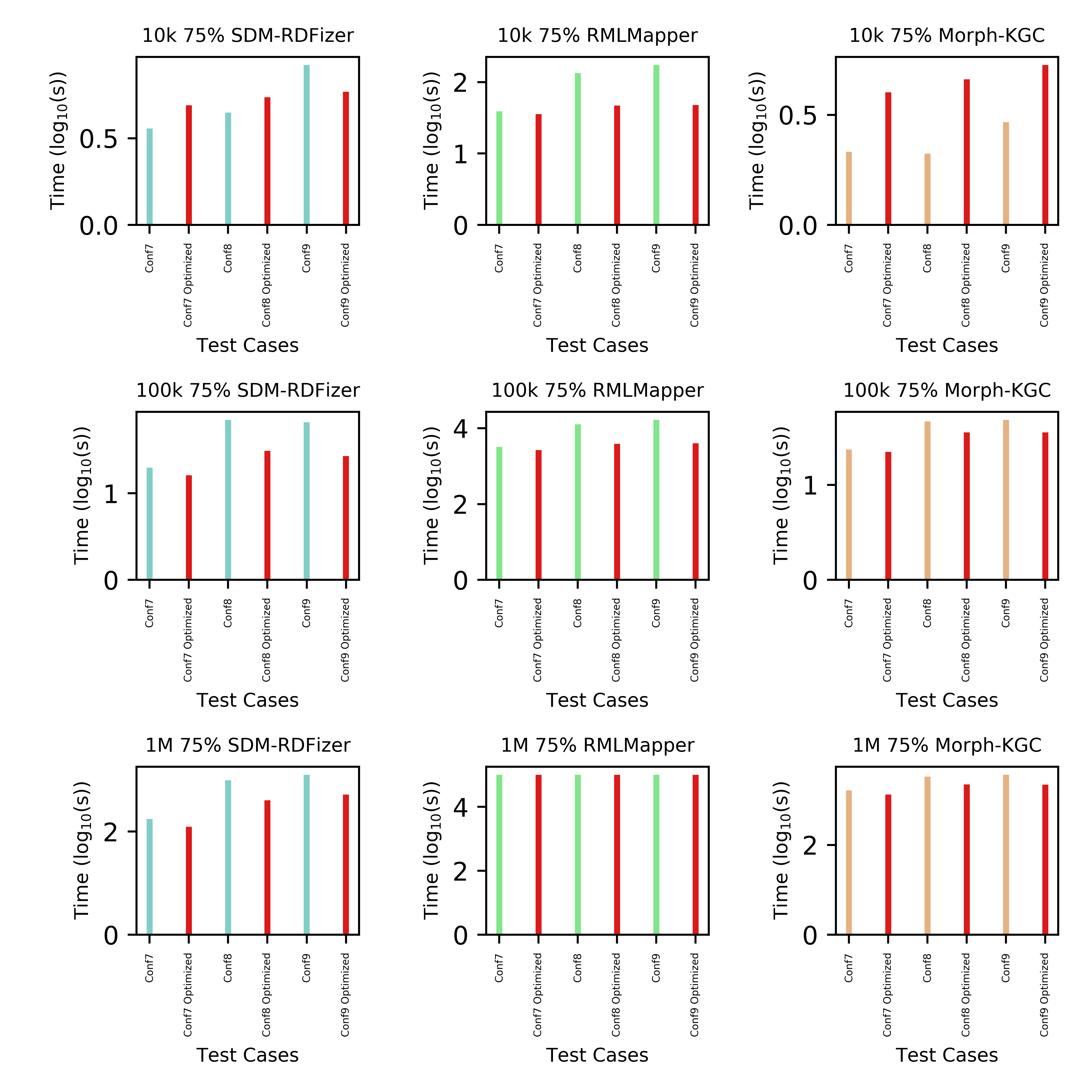}
    \caption{\textbf{Efficiency Planning For Complex Cases with 75\% duplicate rate}. The effects of proposed planning techniques over the SDM-Genomic-Datasets with 75\% duplicate rate over Conf7, Conf8, and Conf9. SDM-RDFizer v3.6+Planning, RMLMapper+Planning, Morph-KGC+Planning}
    \label{fig:complex_75}
\end{figure*}
\noindent
Conversely, in complex cases such as \textbf{Conf6} which include several multi-source role mapping assertions, execution time is reduced significantly exploiting planning. Unfortunately, both RMLMapper and RocketRML lack efficient implementations of the operators that are required to execute referenced-source role mapping assertions. 
\\\noindent
Therefore, the two mentioned engines are unable to finish the execution of \textbf{Conf3} and \textbf{Conf4} before the timeout (i.e., 5 hours). The results in \autoref{fig:genomic} also suggest that with the growth of duplicate data rate, the benefits of using the proposed planning techniques also increased.
\begin{table*}[h!]
\resizebox{\linewidth}{!}{
\begin{tabular}{|l|c|c|c|c|c|c|c|c|c|c|}
\hline

    \multicolumn{11}{|c|}{\cellcolor[HTML]{7C98AA} Percentage of Duplicates: 25\%}\\
    \hline
    \cellcolor[HTML]{AED6F1} Size & \cellcolor[HTML]{AED6F1} Engine  &  \multicolumn{3}{c|}{\cellcolor[HTML]{AED6F1} Conf7} & \multicolumn{3}{c|}{\cellcolor[HTML]{AED6F1} Conf8}  & \multicolumn{3}{c|}{\cellcolor[HTML]{AED6F1} Conf9}\\ 
     \cellcolor[HTML]{AED6F1}&  \cellcolor[HTML]{AED6F1} &  \cellcolor[HTML]{AED6F1}Original & \cellcolor[HTML]{AED6F1}Optimized & \cellcolor[HTML]{AED6F1}\% Savings & \cellcolor[HTML]{AED6F1}Original & \cellcolor[HTML]{AED6F1}Optimized & \cellcolor[HTML]{AED6F1}\% Savings & \cellcolor[HTML]{AED6F1}Original & \cellcolor[HTML]{AED6F1}Optimized & \cellcolor[HTML]{AED6F1}\% Savings\\
\hline 
    \multirow{3}{4em}{ 10k} & SDM-RDFizer  & 3.91 sec & 5.04 sec & -28.90 \% & 5.59 sec & 6.54 sec & -16.99 \%& 10.7 sec & 6.47 sec & 39.53\%\\
    & RMLMapper  & 47.43 sec & 36.69 sec & 22.64 \%&  140.27 sec & 43.93 sec & 68.68 \% & 180.85 sec & 43.25 sec & \textbf{76.09 \%}\\
    &  Morph-KGC & 1.81 sec & 3.55 sec & -96.13\%& 1.79 sec & 4.22 sec & \underline{-135.75} \% & 2.28 sec & 5.2 sec & -128.07 \%\\
\hline
    \multirow{3}{4em}{ 100k} & SDM-RDFizer  & 21.14 sec & 16.88 sec & 20.15 \% & 99.88 sec & 51.11 sec & 48.82 \% & 105.72 sec & 44.97 sec &  57.46 \% \\
    & RMLMapper  & 3205.37 sec & 2628.13 sec &  18.01 \% & 11961.81 sec & 3901.14 sec &  67.38 \%& 12593.16 sec & 3401.17 sec &  \textbf{72.99 \%}\\
     &  Morph-KGC & 20.4 sec & 19.35 sec &  \underline{5.14 \%}& 43.87 sec & 29.38 sec &  33.02 \% & 42.43 sec & 30.84 sec &  27.31 \%\\
\hline
   \multirow{3}{4em}{ 1M}  & SDM-RDFizer  & 177.35 sec & 124.08 sec &  30.03 \% & 1656.29 sec & 607.06 sec & 63.34 \% & 1769.29 sec & 685.22 sec & \textbf{61.27 \%}\\
    & RMLMapper  & TimeOut & TimeOut & - & TimeOut & TimeOut  &- &TimeOut & TimeOut & -\\
    &  Morph-KGC & 1532.94 sec & 1224.37 sec & \underline{20.13 \%} & 3369.11 sec & 2154.92 sec & 36.03 \% & 3329.16 sec & 2071.63 sec & 37.77 \%\\
\hline
    \multicolumn{11}{|c|}{\cellcolor[HTML]{7C98AA} Percentage of Duplicates: 75\%}\\
    \hline
    \cellcolor[HTML]{AED6F1} Size & \cellcolor[HTML]{AED6F1} Engine  &  \multicolumn{3}{c|}{\cellcolor[HTML]{AED6F1} Conf7} & \multicolumn{3}{c|}{\cellcolor[HTML]{AED6F1} Conf8}  & \multicolumn{3}{c|}{\cellcolor[HTML]{AED6F1} Conf9}\\ 
     \cellcolor[HTML]{AED6F1} &  \cellcolor[HTML]{AED6F1} &  \cellcolor[HTML]{AED6F1} Original & \cellcolor[HTML]{AED6F1} Optimized & \cellcolor[HTML]{AED6F1} \%Savings & \cellcolor[HTML]{AED6F1} Original & \cellcolor[HTML]{AED6F1} Optimized & \cellcolor[HTML]{AED6F1} \%Savings & \cellcolor[HTML]{AED6F1} Original & \cellcolor[HTML]{AED6F1} Optimized & \cellcolor[HTML]{AED6F1} \%Savings\\
\hline 
    \multirow{3}{4em}{10k} & SDM-RDFizer  & 3.6 sec & 4.89 sec & -35.83 \% & 4.44 sec & 5.44 sec & -22.52 \% & 8.35 sec & 5.85 sec & 29.94 \% \\
    & RMLMapper  & 38.82 sec & 35.41 sec & 8.78 \% &  133.96 sec & 47.01 sec & 64.90 \% & 173.08 sec & 47.64 sec &\textbf{72.47 \%} \\
    &  Morph-KGC & 2.15 sec & 4.01 sec & -86.51\%  & 2.11 sec & 4.59 sec & \underline{-117.53\%} & 2.93 sec & 5.33 sec & -81.91\%\\
\hline
    \multirow{3}{4em}{100k} & SDM-RDFizer  & 19.72 sec & 16.16 sec &18.05\% & 70.5 sec & 31.06 sec & 55.94\% & 66.15 sec & 29.97 sec & 54.69\%\\
    & RMLMapper  & 3203.19 sec & 2672.59 sec & 16.56\% & 12669.84 sec & 3861.29 sec & 69.52\% & 16541.84 sec & 3985.06 sec & \textbf{75.90\%}\\
     &  Morph-KGC & 23.53 sec & 22.21 sec & \underline{5.60\%} & 46.35 sec & 35.7 sec & 22.97\% & 48.13 sec & 35.68 sec & 25.86\%\\
\hline
   \multirow{3}{4em}{1M}  & SDM-RDFizer  & 174.11 sec & 123.77 sec & 28.91\% & 983.53 sec & 402.59 sec & 59.06\% &  1252.27 sec & 516.99 sec & \textbf{58.71\%}\\
    & RMLMapper  & TimeOut & TimeOut & - & TimeOut & TimeOut & - &TimeOut & TimeOut & - \\
    &  Morph-KGC & 1628.69 sec & 1330.01 sec & \underline{18.33\%} & 3338.93 sec & 2229.78 sec & 33.21\% & 3641.57 sec & 2200.08 sec & 39.58\%\\
\hline
\end{tabular}}
\caption{\textbf{SDM-Genomic-Datasets Complex Test Cases}. Duplicate rates are 25\% and   75\%; Highest Percentage of Savings are highlighted in \textbf{bold}. Lowest Percentage of Savings are \underline{underlined}. The proposed  planning and execution techniques are able to enhance the performance of RMLMapper and speed up execution time by up to 76.08\%; even in the cases, where RMLMapper timed out, the proposed techniques empower RMLMapper to produce intermediate results.
In case of small data sets (e.g., 10K), the proposed techniques may produce overhead in SDM-RDFizer and Morph-KGC (e.g., Conf7 and Conf8). }
\label{table:1}
\end{table*}
\subsection{Experiment 3- Efficiency on Large Datasets}
This experiment evaluates the impact of a data source size on memory usage during the KG creation process. For this purpose, four data sources with different sizes are generated using the GTFS-Madrid-Bench including 1-csv, 5-csv, 10-csv, and 50-csv. Since RMLMapper and RocketRML are not able to scale up to large data sources, we compare the performance of SDM-RDFizer v4.0 in absence and presence of planning; we refer to the latest one as SDM-RDFizer v4.0$^{--}$+Planning. We evaluate the performance of the mentioned versions in terms of both execution time (in second) and main memory consumption (MB); the results of both are reported in log scale. As demonstrated in Figures \ref{fig:experiment3}a and \ref{fig:experiment3}b both versions of SDM-RDFizer are able to complete the KG creation process for all the datasets. Additionally, it can be observed that the planning reduces the memory usage and execution time in each dataset. 
The observed results in Figure \ref{fig:experiment3}a and Figure \ref{fig:experiment3}b suggest that the impact of our proposed planning techniques in the enhancement of the performance of SDM-RDFizer v4.0$^{--}$ is higher than the planning techniques implemented by SDM-RDFizer v4.0.
\subsection{Experiment 4- Efficiency on  Complex Mappings}
This experiment aims at assessing the effect of the complex mapping assertions on the execution time during the KG creation process. In these experiments, RocketRML is replaced by Morph-KGC since RocketRML is unable to execute the multi-source mapping assertions that composed the \textbf{Conf7}, \textbf{Conf8}, and \textbf{Conf9}. 
\\\noindent
Figures \ref{fig:complex_25} and \ref{fig:complex_75} report on execution time (log scale) and Table \ref{table:1} presents the specific values of each execution.
As observed, the RMLMapper performance is improved in \textbf{Conf7}, \textbf{Conf8}, and \textbf{Conf9} even in data sources of small size, i.e., 10k.
In the data source of the size 10k, there is $22.64\%$ reduction of execution time for \textbf{Conf7} with $25\%$ duplicate rate and $8.78\%$ reduction with $75\%$ duplicate rate, $68.68\%$ reduction for \textbf{Conf8} with $25\%$ duplicate rate and $64.9\%$ reduction with $75\%$ duplicate rate, and $76.09\%$ reduction for \textbf{Conf9} with $25\%$ duplicate rate and $72.47\%$ reduction with $75\%$ duplicate rate. For 100k, there is a $18.01\%$ reduction of execution time for \textbf{Conf7} with $25\%$ duplicate rate and $16.56\%$ reduction with $75\%$ duplicate rate, a $67.38\%$ reduction for \textbf{Conf8} with $25\%$ duplicate rate and $69.52\%$ reduction with $75\%$ duplicate rate, and a $72.99\%$ reduction for \textbf{Conf9} with $25\%$ duplicate rate and $75.90\%$ reduction with $75\%$ duplicate rate. 
\\\noindent
The RMLMapper timed out after 5 hours with both methods when executing the 1M data sources with all three mappings with duplicate rates. This can be attributed to how the execution of the join is implemented in the RMLMapper and the size of the data. But with the planned execution, it could generate at least a portion of the KG for each mapping. For \textbf{Conf7}, \textbf{Conf8}, and  \textbf{Conf9}, respectively,  $32.65\%$, $24.82\%$, and $28.69\%$ of the KG are generated. 
\\\noindent
For the SDM-RDFizer and Morph-KGC, there was overhead when generating the KG for \textbf{Conf7} and \textbf{Conf8} with 10k. This can be attributed to the fact that both the SDM-RDFizer and Morph-KGC already have optimization techniques implemented. Combining the optimization techniques and the physical plan causes the overhead in cases with small data sources, i.e., 10k. While for \textbf{Conf9}, there is a $39.53\%$ reduction with $25\%$ duplicate rate and a $29.94\%$ reduction with $75\%$ duplicate rate for the SDM-RDFizer when using the planned execution. There are savings of 100k and 1M when using the planned execution for both engines. In particular, \textbf{Conf9} presents the highest savings. For 100k, there is a $57.46\%$ reduction with $25\%$ duplicate rate and a $54.69\%$ reduction with $75\%$ duplicate rate for the SDM-RDFizer and a $27.31\%$ reduction with $25\%$ duplicate rate and a $25.86\%$ reduction with $75\%$ duplicate rate for Morph-KGC. 
\\\noindent
For 1M, there is a $61.27\%$ reduction with $25\%$ duplicate rate and a $58.71\%$ reduction with $75\%$ duplicate rate for the SDM-RDFizer and a $37.77\%$ reduction with $25\%$ duplicate rate and a $39.58\%$ reduction with $75\%$ duplicate rate for Morph-KGC. This increase in savings is related to the complexity of the mapping; higher complexity causes higher savings.
\\\noindent
In conclusion, applying the proposed planning techniques reduces the execution time, independent of the engine by which they are adopted. However, applying these techniques in engines such as SDM-RDFizer and Morph-KGC, which already perform optimization techniques, may cause an overhead. 
Specifically, in the case of having small size data sources or less complex mapping assertions, the cost of planning in addition to the other optimization techniques implemented in the engine can be higher than the savings. Like any optimization technique, there is a trade-off that can be estimated based on the provided data integration system. The higher the complexity of the mapping assertions and dataset size, the higher the execution time improvement.  
\subsection{Discussion}
\noindent\textbf{Answer to RQ1.} There exist configurations of data integration systems where the proposed planning techniques improve the performance of any state-of-the-art engines.  The experimental results provide insights on the cases where planning improves the KG creation frameworks in contrast to the ones that it may cause negative impact. E.g., in case of having small data sources or simple mapping assertions, the execution times of SDM-RDFizer and Morph-KGC are lower ignoring the planning of the mapping assertions. However, it is important to note that execution planning empowers state-of-the-art engines without continuous behavior to generate a partial KG output. In other words, the generated plans enable some engines to produce outputs instead of timing out or running out of memory.
\\\noindent\textbf{Answer to RQ2.} Attribute mapping assertion presents the shortest execution time of all the types of mapping assertion since they represent a simple projection of the raw data. In terms of memory usage, attribute mapping assertion dependent on the size of the data source, meaning larger data sources cause greater memory usage. The execution time of a multi-source role mapping assertion depends on the size of the data sources and the number of values associated with them. Larger data sources and a more significant number of associated values imply higher memory usage. The execution time of referenced-source role mapping assertions depends on the size of the data source and the data management techniques implemented for each engine. RMLMapper and RocketRML execute the mentioned operation as a Cartesian product, causing the execution time to grow exponentially and, by extension, the memory usage.  
\\\noindent\textbf{Answer to RQ3.} Algorithm \ref{alg:bushytree} generates a bushy tree, which schedules which mapping assertions should be executed together because of the number of predicates or data sources in common. Executing  mapping assertions following a bushy tree plan reduces both execution time and memory usage. In attribute mapping assertions with the same data source or referenced source, role mapping assertions have minimal impact on execution time and memory usage. Since all mapping assertions in question use the same data source, only one partition would be used. For multi-source role mapping assertion, Algorithm \ref{alg:bushytree} generates bushy trees whose execution positively influences time and memory. This behavior is achieved by partitioning mapping assertions that reduce the number of operations per group. Therefore, the workload, execution time, and memory usage are reduced.
%%%%%%%%%%%%%%%%%%%%%%%%%%%%%%%%%%%%%%%%
%%            CONCLUSIONS             %%
%%%%%%%%%%%%%%%%%%%%%%%%%%%%%%%%%%%%%%%%
\section{Conclusions and Future Work}
\label{sec:conclusions}
We address the problem of efficient KG creation. This problem is of paramount relevance given the momentum that KGs have gained in science and industry, as well as declarative processes to specify KGs. 
We present heuristic-based solutions that, following greedy algorithms, can identify execution plans that can efficiently generate KGs. The empirical evaluation of the proposed methods empowers existing RML-compliant engines and enables them to scale to complex situations. The execution planning techniques partition mapping assertions and schedule them into execution plans that consume less memory and reduce execution time. Thus, the proposed planning methods evidence the crucial role that optimization techniques-- defined in the context of query processing-- also have in the KG creation process. 
Moreover, the reported results put in perspective the need of specialized data management methods for scaling up KG creation to complex data integration systems present in real-world applications.
Albeit efficiently defined, execution planning may be costly and generate overhead, which negatively impact engine behavior in simple cases. In the future, we will research lightweight cost-based planning methods to estimate more efficient execution schedulers.  

%%%%%%%%%%%%%%%%%%%%%%%%%%%%%%%%%%%%%%%%
%%          ACKNOWLEDGEMENTS          %%
%%%%%%%%%%%%%%%%%%%%%%%%%%%%%%%%%%%%%%%%
\section*{Acknowledgements}
This work has been partially supported by the EU H2020 RIA
funded project CLARIFY with grant agreement No 875160 and PLATOON (GA No. 872592). Federal Ministry for Economic Affairs and Energy of Germany in the project CoyPu (project number 01MK21007[A-L]. Furthermore, Maria-Esther Vidal is partially supported by the Leibniz Association in the program "Leibniz Best Minds: Programme for Women Professors", project TrustKG-Transforming Data in Trustable Insights with grant P99/2020.

%%%%%%%%%%%%%%%%%%%%%%%%%%%%%%%%%%%%%%%%
%%             REFERENCES             %%
%%%%%%%%%%%%%%%%%%%%%%%%%%%%%%%%%%%%%%%%

\bibliographystyle{abbrv}
\bibliography{bibliography}

\appendix
\section{Theorems and Proofs}
\label{sec:optimality}

\subsection{Theorem of Optimality}
\noindent \autoref{theo:optimality}. 
Let $DIS_\mathcal{G}=\langle O,S,M \rangle$ be a data integration system such that assertions in $M$ meet the following conditions:
\begin{itemize}
\item A concept mapping assertion $ma_j$ on source $S_j$ is referred from any number of multi-source role mapping assertions $ma_i$, but these assertions are all from one source $S_i$. 
\item A property $p$ from $O$ is defined, at most, on one mapping assertion $ma_i$. 
\end{itemize}
Let $BT$ be a bushy tree plan over mapping assertions in $M$ and data sources in $S$; $BT$ generates $G$ and respects the optimality principles \textbf{P1}-\textbf{P4}. Then, $BT$ is optimal, i.e., there is no other equivalent bushy tree plan $BT'$ such as $fu(BT',S) < fu(BT,S)$.
\\\noindent \textbf{Proof}. By contradiction. Assume $BT$ respects the optimality principles \textbf{P1}-\textbf{P4}, but there is a different bushy tree plan $BT'$, i.e., the executions of $BT$ and  $BT'$ produce the same RDF triples when evaluated on the same engine, and $fu(BT',S)<fu(BT,S)$. 
\\\noindent \textbf{Base Case}
Suppose $BT$ is an intra-source partition $G_k$ that includes all the mapping assertions over a source $S_i$, i.e., concept, attribute, and single-source and referenced-source role mapping assertions. $BT'$ is an equivalent bushy tree plan, but it is different from $BT$. It comprises at least two intra-source partitions for the mapping assertions in $G_k$. These partitions contain only assertions over $S_i$; for hypothesis, a predicate can be defined by at most one mapping assertion. However, this would lead to a contradiction because the execution of $BT'$ will require loading in memory $S_i$ several times, during the evaluation of the two intra-source partitions and $fu(BT,S) \leq fu(BT',S)$.
\\\noindent Suppose $BT$ is an inter-source partition $G_k$ over sources $S_i$ and $S_j$, which are related via multi-source role mapping assertions.
$G_k$ includes all the multi-source role mapping assertions from $S_i$ to the concept mapping assertion over $S_j$, and all the attribute, single-source, and referenced-source role mapping assertions over $S_j$. Note that by hypothesis, the concept mapping assertion over $S_j$ is referred from only multi-source role mapping assertions over $S_i$. 
An equivalent plan $BT'$ should have at least two partitions, i.e., $G_{k,1}$ and $G_{k,2}$. Without lost of generality, assume that $G_{k,1}$ is an intra-source partition over $S_j$, while $G_{k,2}$ comprises the multi-source role mapping assertions on $S_i$ that refer to $S_j$. $G_{k,1}$ and $G_{k,2}$ collect data from $S_j$, and both upload $S_j$ independently. Thus, $fu(BT,S) \leq fu(BT',S)$, leading, thus, a contradiction. 
\\\noindent \textbf{Inductive Hypothesis}
 $BT_1$ and $BT_2$ are optimal bushy plans for $M$. 
\\\noindent \textbf{Induction Step}
Suppose $BT$ comprises sub-plans $BT_1$ and $BT_2$ and the union operator $OP$. Let $BT'$ an equivalent bushy tree plan. Without lost of generality, assume $BT'$ implements the eager evaluation of $DR$ operators to eliminate duplicates of a property $p$, while $BT$ follows a lazy evaluation. If $DR$ is required, the instances of $p$ are generated during the execution of at least two partitions.  
However, this leads to a contraction, since by hypothesis, every predicate $p$ is defined by at most one mapping assertion, which should be included in only one partition, either in $BT_1$ or $BT_2$, because both are optimal. \qed

\subsection{Time Complexity of Algorithm \ref{alg:bushytree}}
\noindent \autoref {theo:complexity}. Let $G^{GP_{M}}$ be a graph plan of the groups in $GP_{M}$. Let $n$ be the $G^{GP_{M}}$ cardinality, i.e., the number of groups in $GP_{M}$. The time complexity of Algorithm \ref{alg:bushytree} is $O(n \log n)$ and up to $2^n -1$ bushy sub-plans are generated. 

\noindent \textbf{Proof}.
Algorithm \ref{alg:bushytree} traverses the space of bushy tree plans in iterations until a fixed point on hyper-graph is reached. Initially, the $n$ partitions (i.e., nodes) are sorted based on degree in the graph plan $G^{GP_{M}}$ and number of shared properties; this is done in $O(n \log n)$.  
Then, in the first iteration, $n$ hyper-nodes are generated in the hyper-graph, each one composes one group in $GP_{M}$. Next, the nodes are visited in the identified order, and $\lceil \frac{k}{2} \rceil$ hyper-nodes are created. This process continues until iteration $\lceil \log_2 n \rceil$. In total, up to $2^n -1$ hyper-nodes are generated. These hyper-nodes correspond to bushy sub-plans.
\qed

\end{document}